\documentclass{article}

\PassOptionsToPackage{numbers,compress}{natbib}
\usepackage[preprint]{neurips_2026}

\usepackage[utf8]{inputenc}
\usepackage[T1]{fontenc}
\usepackage{hyperref}
\usepackage{url}
\usepackage{booktabs}
\usepackage{amsfonts}
\usepackage{amsmath, amssymb}
\usepackage{nicefrac}
\usepackage{microtype}
\usepackage{xcolor}
\usepackage[table]{xcolor}
\usepackage{graphicx}
\usepackage{multirow}
\usepackage{makecell}
\usepackage{subcaption}
\usepackage{array}
\usepackage{makecell}
\usepackage{wrapfig}
\usepackage{listings}
\usepackage{algorithm}
\usepackage{algpseudocode}
\usepackage{threeparttable}
\usepackage{booktabs}
\usepackage{tabularx}
\newcolumntype{Y}{>{\centering\arraybackslash}X}

\usepackage{xspace}

\newcommand{\modelname}{\textsc{CAVE}\xspace}
\newcommand{\datasetname}{{TRACER-Bench}\xspace}

\title{CAVE: A Structured Credit Assignment Approach for Fragmented Visual Evidence Reasoning}

\author{%
\begin{minipage}{\textwidth}
\centering
Tengda Guo$^{1,*}$ \quad
Jie Leng$^{1,*}$ \quad
Hanlei Li$^{1,*}$ \quad
Yaoyuan Liang$^{1,\dagger}$ \quad
Qingyue Zhang$^1$\\
Dian Yang$^1$ \quad
Mingyu Zhang$^2$ \quad
Yuhua Fu$^3$ \quad
Shao-Lun Huang$^{1,\dagger}$\\[0.5em]
$^1$Tsinghua University \quad
$^2$Peking University \quad
$^3$Zhejiang University of Technology\\
$^*$Equal contribution. \quad
$^\dagger$Corresponding author.
\end{minipage}
}

\begin{document}

\maketitle

\begin{abstract}
\label{sec:abstract}
Vision-Language Models (VLMs) have achieved strong performance on general multimodal reasoning, yet remain challenged in integrating nonlocal visual information to support semantically underdetermined visual reasoning. We describe this challenge as Fragmented Visual Reasoning. To this end, we propose Credit Assignment for Visual Evidence (CAVE), a structured process-reward method based on GRPO for interleaved visual reasoning. Specifically, CAVE evaluates the contribution of intermediate steps at the action level via three complementary reasoning process signals: belief update, evidence acquisition, and adaptive focus control, thereby guiding the model to optimize each reasoning action and learn more reliable visual reasoning strategies. Meanwhile, we construct TRACER-Bench, which covers four nonlocal and semantically confusable reasoning dimensions and provides key intermediate evidence to supervise reasoning paths. Experiments demonstrate that CAVE substantially improves performance on tasks requiring fragmented visual evidence integration, covering both public benchmarks and our newly introduced TRACER-Bench, while retaining competitive performance on general multimodal evaluations. Further analyses reveal that CAVE effectively improves the visual reasoning capacity and exhibits stronger robustness under longer-range and deeper cross-region dependencies.

\end{abstract}

\section{Introduction}
\label{sec:introduction}
\vspace{-10pt}

\begin{figure}[t]
    \centering
    \includegraphics[width=\linewidth]{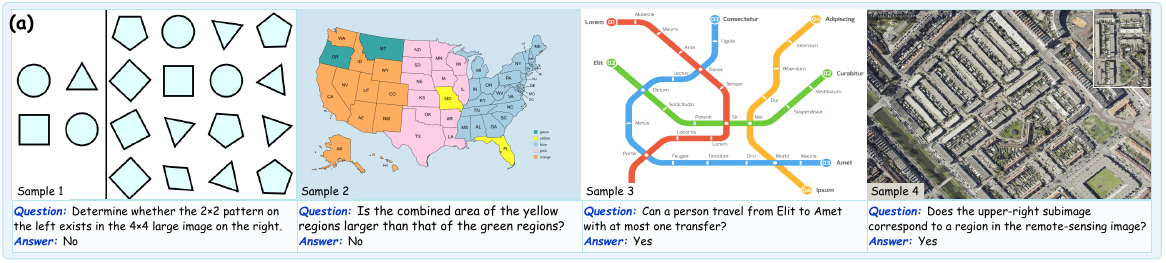}
    \vspace{-2pt}
    \includegraphics[width=\linewidth]{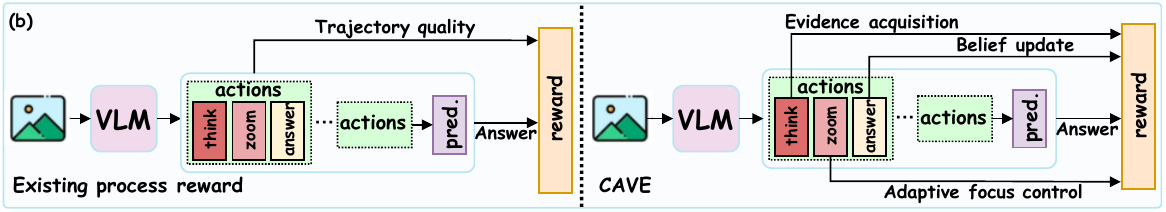}
    \vspace{-10pt}
    \caption{(a) Samples of Fragmented Visual Reasoning. The examples cover Geometry, Geography, Route, and Remote-Sensing tasks. Samples 1--3 require organizing confusable and subtle evidence across separate regions, while Sample 4 requires matching nonlocal relations in a noisier real image. (b) Comparison between existing process-reward methods and CAVE. Existing methods usually rely on trajectory-level or answer-side optimization, whereas CAVE assigns action-level structured credits according to the reasoning contributions of intermediate visual-evidence reasoning steps.}
    \label{fig:tracer_examples}
    \vspace{-15pt}
\end{figure}

Vision-Language Models (VLMs) have made substantial progress in multimodal reasoning tasks \citep{alayrac2022flamingo,li2023blip2,liu2023llava,bai2025qwen25vl}, largely benefiting from the semantic priors accumulated through large-scale pretraining and the chain-of-thought capabilities of language models \citep{wei2022chain,zhang2023multimodalcot}. However, when tasks primarily rely on basic visual capabilities with low language priors, existing VLMs still exhibit clear brittleness. 
Representative tasks in Tunnel Vision \citep{berman2025tunnelvision,chen2026babyvision} provide a concrete example: even simple human-solvable problems can require chaining evidence across multiple, potentially distant image regions, where current VLMs perform poorly. Building on this observation, we study a broader setting in which task-critical evidence is spatially fragmented and semantically weakly separable across regions. We refer to this challenge as Fragmented Visual Reasoning (FVR). As illustrated in Figure~\ref{fig:tracer_examples}(a), FVR creates a reasoning dilemma: if VLMs rely only on coarse global visual perception, their vision-side representations often fail to robustly capture long-range dependencies and subtle local differences, thereby biasing reasoning paths toward a verbalized subset of visual evidence. As a result, models may produce coherent but visually ungrounded reasoning chains, leading to error propagation in subsequent reasoning steps.


This challenge naturally connects to recent interleaved multimodal reasoning methods, which introduce a ``Think--with--Image'' paradigm, enabling models to progressively acquire visual information through localized re-observation and providing a more natural solution form for such tasks \citep{hu2024visualsketchpad,su2025openthinkimg,zheng2025deepeyes,zhang2025chainoffocus}. However, even though existing process-reward methods attempt to alleviate sparse feedback in long-trajectory training \citep{wang2025visualprm,wang2026igpo}, they usually evaluate intermediate steps from the perspective of final-answer improvement or overall trajectory quality, failing to provide action-specific process supervision and thus leaving models without targeted guidance to learn reliable strategies for each distinct reasoning behavior in visual evidence exploration, as shown in Figure~\ref{fig:tracer_examples}(b) left.

To address this, we propose Credit Assignment for Visual Evidence (CAVE), which performs action-level structured credit assignment over multi-round reasoning trajectories for visual evidence exploration~\citep{shao2024deepseekmath}, as shown in Figure~\ref{fig:tracer_examples}(b) right. \modelname evaluates intermediate reasoning with three complementary signals: belief update, evidence acquisition, and adaptive focus control. These signals respectively assess progress toward the correct answer, recovery of task-critical visual details, and the appropriateness of visual region focusing, thereby explicitly guiding the model to learn more reliable evidence discovery, reasoning state updating, and goal-directed exploration behaviors. To enable controlled evaluation and process-level analysis, we construct \datasetname, which covers three abstract scenarios and one real-world scenario: Rule-Switching Navigation, Nonsemantic Tracing, Embedded Matching, and Remote-Sensing Subimage Matching. These scenarios require cross-regional evidence with low semantic separability, and \datasetname provides corresponding key evidence for offline analysis.

In summary, our main contributions are as follows: 1) We introduce Fragmented Visual Reasoning, a challenging task characterized by low semantic separability across multiple visual regions, together with TRACER-Bench for systematic evaluation across diverse scenarios with verifiable explicit visual-evidence chains. 2) We propose {CAVE}, a structured credit assignment process-reward method based on GRPO for interleaved visual reasoning. 3) Experiments show that CAVE effectively improves model performance on benchmarks with fragmented vision, achieves competitive results on general multimodal evaluations, and effectively improves the visual reasoning capacity.

\vspace{-10pt}
\section{Related work}
\vspace{-10pt}

\noindent\textbf{VLMs visual reasoning and interleaved visual reasoning.} VLMs have achieved significant progress on general multimodal reasoning tasks. Representative works such as Flamingo and LLaVA have advanced the development of general-purpose VLMs through interleaved image-text modeling and related techniques \citep{alayrac2022flamingo,li2023blip2,liu2023llava,bai2025qwen25vl,chen2024internvl,li2024llavaonevision}. However, existing VLMs remain fragile in tasks requiring 
detailed discrimination and nonlocal evidence. Tunnel Vision, BabyVision, and related benchmarks reveal this issue from the perspectives of long-range region association and basic visual cognition, respectively \citep{berman2025tunnelvision,chen2026babyvision,zhang2025mmerealworld,fu2024blink,guan2024hallusionbench}. To improve reasoning capability, methods such as DDCoT and LLaVA-CoT introduce staged or visually grounded intermediate reasoning \citep{zheng2023ddcot,xu2024llavacot,zhang2023multimodalcot,ni2025pointrft,wu2025groundedcot}. Furthermore, Visual Sketchpad, VisProg, ViperGPT, DeepEyes, and related works extend VLMs reasoning into a multi-round ``reasoning--observing--reasoning'' process through tool use or adaptive zooming \citep{hu2024visualsketchpad,gupta2023visprog,suris2023vipergpt,su2025openthinkimg,zheng2025deepeyes,zhang2025chainoffocus}. However, these methods mainly focus on intermediate reasoning or local observation mechanisms, while still lacking dedicated modeling of how observations serve as accumulated visual states for subsequent cross-region dependency progression.

\noindent\textbf{GRPO and process rewards.} GRPO/R1-style reinforcement learning shows that rule-based or final-answer rewards can elicit long-chain reasoning capabilities in large models \citep{shao2024deepseekmath,guo2025deepseekr1}, and has been extended to Vision-Language Models by works such as VLM-R1, R1-VL, and OpenVLThinker \citep{shen2025vlmr1,zhang2025r1vl,deng2025openvlthinker}. However, existing GRPO-style VLMs training usually relies on final answers, format constraints, or general rule-based rewards, making it difficult to characterize the process-level contributions of observation, evidence acquisition, and region transition in multi-round visual reasoning. To alleviate reward sparsity in long-trajectory training, VisualPRM, StepSearch, GiGPO, and IGPO explore process rewards, step-level feedback, trajectory comparison, and information-gain signals \citep{wang2025visualprm,zheng-etal-2025-stepsearch,feng2025gigpo,wang2026igpo}. However, such process signals usually focus more on the overall contribution of states to final task utility or answer formation, and do not explicitly distinguish different types of evidence progression in multi-round visual reasoning. In contrast, CAVE targets visual evidence progression and structurally decomposes intermediate contributions into three types of process credit: belief update, evidence acquisition, and adaptive focus control.



\vspace{-10pt}
\section{Method}
\label{sec:method}
\vspace{-10pt}

\begin{figure}[t]
    \centering
    \includegraphics[width=\linewidth]{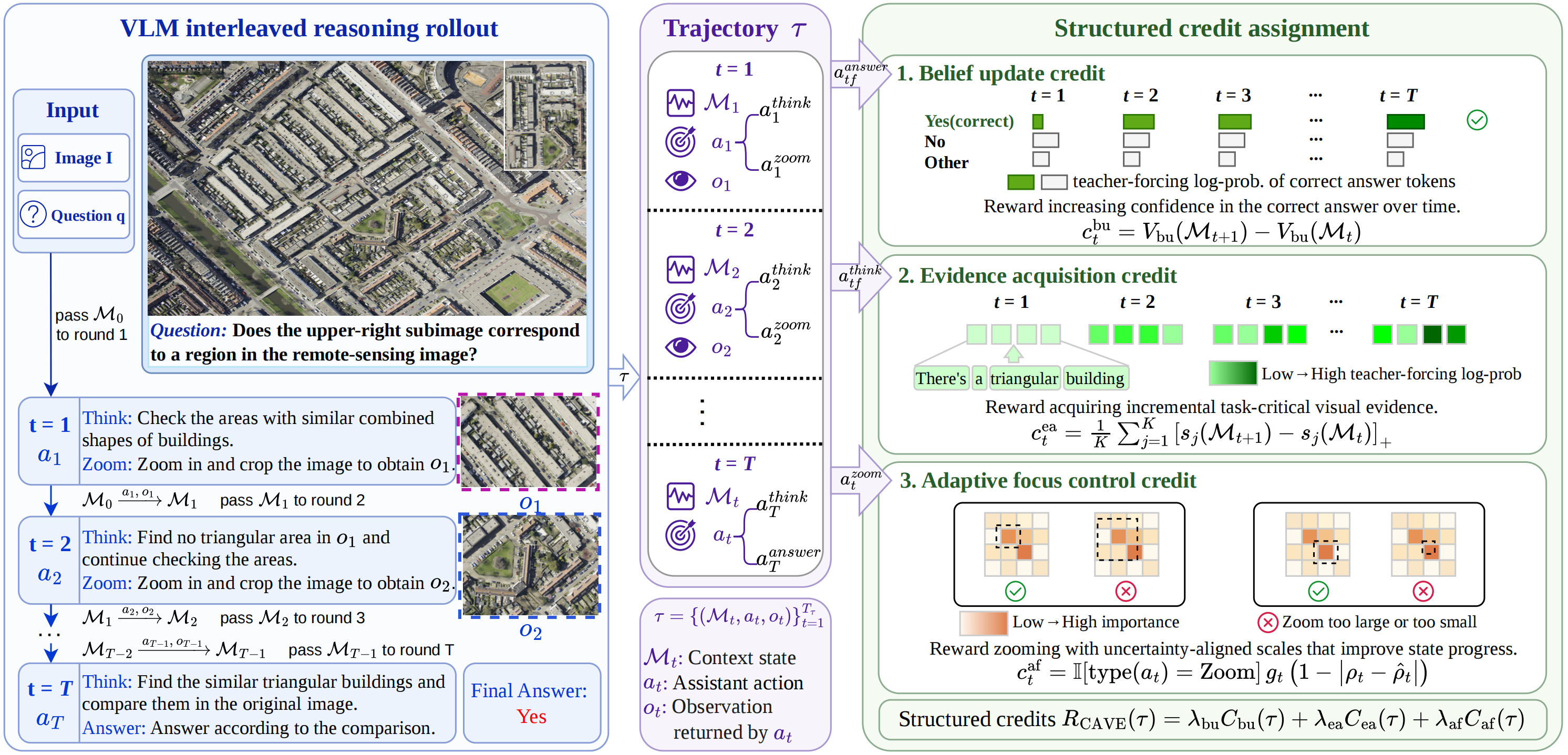}
    \vspace{-4pt}
    \caption{Overview of \modelname. Given an input pair, the policy performs fragmented visual evidence reasoning. CAVE assigns structured process credits to belief update, evidence acquisition, and adaptive focus control, and is optimized with GRPO. $a_{\mathrm{tf}}$ denotes the teacher-forced action under the current context state, obtained outside the rollout and used only for reward computation.}
    \label{fig:cave_framework}
    \vspace{-12pt}
\end{figure}

The goal of \modelname is to optimize the model's capability of fragmented visual evidence reasoning during interleaved visual reasoning. Different from measuring intermediate states with a single information gain or final task utility, \modelname focuses on whether intermediate actions promote answer-belief update, key evidence acquisition, and adaptive focus control. To this end, we first define the model's interleaved reasoning trajectory, then introduce three types of structured process credits, and describe how they are integrated into GRPO optimization \citep{shao2024deepseekmath}, with details shown in Figure~\ref{fig:cave_framework}.

\vspace{-8pt}
\subsection{Interleaved visual reasoning trajectory}
\vspace{-5pt}

Given an input $x=(I,q)$, where $I$ denotes the image and $q$ denotes the question, \modelname represents the model's solving process as an interleaved visual reasoning trajectory with at most $T$ rounds:

\vspace{-12pt}

\begin{equation*}
\begin{aligned}
\tau=\{(\mathcal M_t,a_t,o_t)\}_{t=1}^{T_\tau},\quad
a_t\sim\pi_\theta(\cdot\mid\mathcal M_t),\quad
\mathcal M_{t+1}=\mathrm{Update}(\mathcal M_t,a_t,o_t),\quad T_\tau\le T .
\end{aligned}
\end{equation*}

\vspace{-6pt}

Here, $\mathcal M_t$ denotes the multi-modal context state at the beginning of round $t$, which contains the original image, the question, historical reasoning texts, and the acquired local visual observations; $a_t$ denotes the assistant action generated by the policy model under the current state, which may correspond to textual reasoning, a zoom-tool call, or the final answer generation; and $o_t$ denotes the observation returned by the environment according to $a_t$, such as the local sub-image after zooming and its additional prompt. This trajectory formulation identifies the components of interleaved reasoning and makes explicit the boundary between policy-generated actions and environment-provided observations. We specify the assistant-token optimization and the associated response-mask setting in the policy optimization section.

\vspace{-8pt}
\subsection{Structured credits for fragmented visual evidence reasoning}
\vspace{-5pt}

During reasoning, the value of an effective intermediate step does not always immediately manifest as an increase in answer confidence. Intermediate steps can serve distinct functional roles: acquiring precursor visual evidence, updating intermediate states, or determining the location and scale of subsequent observation. Based on this, \modelname does not adopt a single state-utility gain, but attributes the state transition $\mathcal M_t \to \mathcal M_{t+1}$ induced by action $a_t$ from three complementary perspectives: answer belief improvement, visual evidence observation, and region focus matching. These credits constitute process-level constraints on fragmented visual evidence reasoning, rather than collapsing intermediate steps into a single undifferentiated state-utility reward signal.

\vspace{-10pt}
\subsubsection{Belief update credit}
\vspace{-5pt}

Belief update credit characterizes answer-side reasoning progress. If a reasoning step or tool call effectively updates the context state, then under the updated $\mathcal M_{t+1}$, the ground-truth answer should have higher conditional log-likelihood. Let $\mathcal A$ denote the token interval of the answer body. We use teacher forcing to define the answer-side state value $V_{\mathrm{bu}}$ and its adjacent-state gain $c_t^{\mathrm{bu}}$ \citep{wang2026igpo}:

\vspace{-12pt}

\begin{equation*}
\begin{aligned}
V_{\mathrm{bu}}(\mathcal M_t)
=
\frac{1}{|\mathcal A|}
\sum_{i\in\mathcal A}
\log p_\theta
\left(
y_i^\star \mid \mathcal M_t, y_{<i}^\star
\right),
\qquad
c_t^{\mathrm{bu}}
=
V_{\mathrm{bu}}(\mathcal M_{t+1})
-
V_{\mathrm{bu}}(\mathcal M_t).
\end{aligned}
\end{equation*}

\vspace{-8pt}

We only score the answer body to avoid the influence of format templates or irrelevant prefixes and suffixes on the estimation. This credit term rewards intermediate steps that move the model state closer to the correct answer, thereby constraining answer-oriented state updates.

\vspace{-8pt}
\subsubsection{Evidence acquisition credit}
\vspace{-5pt}

Relying solely on answer-belief gain is insufficient for multi-round visual reasoning, since an intermediate action may make critical visual clues available before these clues are fully integrated into an answer-side belief. To capture such precursor progress that may be missed by answer-belief gain, we further introduce evidence acquisition credit. Teacher forcing details are in Appendix~\ref{app:teacher_forcing}.

For training samples equipped with task-critical evidence annotations, we represent the intermediate visual information required for answer formation as a set of $K$ textualized visual evidence units, $\mathcal P=\{p_1,\dots,p_K\}$, where each $p_i$ denotes an individual evidence unit, as illustrated by the example ``There's a triangular building'' in Figure~\ref{fig:cave_framework}. These units correspond to necessary visual intermediate information, such as perceptual details, local attributes, cross-region correspondences, path continuity, or structural cues. They are automatically constructed and further verified through human inspection, with detailed construction and quality-control procedures provided in Appendix~\ref{app:quality_control}.

For the $j$-th evidence unit $p_j$, let its token interval be $\mathcal E_j$. We define its recoverability score $s_j$ under state $\mathcal M_t$ and its positive gain as:

\vspace{-22pt}

{
\begin{equation*}
\begin{aligned}
s_j(\mathcal M_t)
=
\frac{1}{|\mathcal E_j|}
\sum_{i\in\mathcal E_j}
\log p_\theta
\left(
p_{j,i}
\mid
\mathcal M_t,\,
p_{j,<i}
\right),
\qquad
c_t^{\mathrm{ea}}
=
\frac{1}{K}
\sum_{j=1}^{K}
\left[
s_j(\mathcal M_{t+1})
-
s_j(\mathcal M_t)
\right]_+ .
\end{aligned}
\end{equation*}
}

\vspace{-13pt}

Here, $[\cdot]_+$ indicates that only positive evidence gains are retained. This term does not require the model to generate standardized explanations during reasoning, but instead transforms whether the current state is closer to the task-critical visual evidence into an optimizable process signal, thereby distinguishing trajectories that blindly guess from those that progressively acquire evidence.

\vspace{-8pt}
\subsubsection{Adaptive focus control credit}
\vspace{-5pt}

Adaptive focus control credit constrains the zoom-based visual 
exploration in interleaved reasoning. An effective zoom should balance evidence coverage and visual noise, providing informative local perception while preserving necessary contextual cues. Rather than merely rewarding the occurrence of a zoom call \citep{zheng2025deepeyes,zhang2025chainoffocus}, CAVE encourages the model to select local observations that both induce state progress and match the current uncertainty.

Specifically, we estimate the uncertainty of the current state using the top-$k$ distribution entropy of answer tokens during answer teacher forcing, denoted as $H_t=\frac{1}{|\mathcal A|}\sum_{i\in\mathcal A}\mathrm{Entropy}_k(i)$, and normalize it to obtain $u_t\in[0,1]$. For a valid zoom at round $t$, let the cropped-region area ratio be $\rho_t$, and define its uncertainty-aware target scale as $\hat{\rho}_t=\rho_{\min}+(\rho_{\max}-\rho_{\min})u_t$.

To avoid rewarding zoom actions with appropriate scale but no actual contribution, we further gate the reward using positive state progress after observation:

\vspace{-13pt}

\begin{equation*}
\begin{aligned}
g_t=
1-\exp\left(-\left([c_t^{\mathrm{bu}}]_+ +  c_t^{\mathrm{ea}}\right)\right),
\qquad
c_t^{\mathrm{af}}
=
\mathbb{I}[\mathrm{type}(a_t)=\mathrm{Zoom}]
\, g_t
\left(1-\left|\rho_t-\hat{\rho}_t\right|\right).
\end{aligned}
\end{equation*}

\vspace{-6pt}

Here, $c_t^{\mathrm{bu}}$ and $c_t^{\mathrm{ea}}$ denote belief update and evidence acquisition credit, respectively. This term does not rely on region-level ground-truth annotations; instead, it rewards a zoom action only if it yields positive belief-update or evidence-acquisition progress, while keeping its 
observation scale aligned with the current uncertainty. Therefore, it provides an adaptive zoom-scale constraint for acquiring fragmented visual evidence.

\vspace{-10pt}
\subsection{Policy optimization with structured credits}
\vspace{-5pt}

\setlength{\textfloatsep}{0pt plus 1pt minus 2pt}

\begin{algorithm}[t]
\caption{Structured credit computation in CAVE}
\label{alg:cave_credit}
\small
\begin{algorithmic}[1]
\Require $\tau=\{(\mathcal M_t,a_t,o_t)\}_{t=1}^{T_\tau}$, answer span $\mathcal A$, evidence units $\mathcal P$
\Ensure $R_{\mathrm{CAVE}}(\tau)$

\For{$t=1$ to $T_\tau$}
    \State $c_t^{\mathrm{bu}} \gets V_{\mathrm{bu}}(\mathcal M_{t+1})-V_{\mathrm{bu}}(\mathcal M_t)$
    \State $c_t^{\mathrm{ea}} \gets \frac{1}{K}\sum_{j=1}^{K}[s_j(\mathcal M_{t+1})-s_j(\mathcal M_t)]_+$
    \State $\hat{\rho}_t \gets \rho_{\min}+(\rho_{\max}-\rho_{\min})\mathrm{NormEntropy}_k(\mathcal A\mid \mathcal M_t)$
    \State $g_t \gets 1-\exp\!\left(-\left([c_t^{\mathrm{bu}}]_+ + c_t^{\mathrm{ea}}\right)\right)$
    \State $c_t^{\mathrm{af}} \gets \mathbb{I}[a_t=\mathrm{Zoom}]\, g_t \left(1-\left|\rho_t-\hat{\rho}_t\right|\right)$
\EndFor

\State $C_m(\tau)\gets\sum_{t=1}^{T_\tau} w_t^m\phi_m(c_t^m),\quad m\in\{\mathrm{bu},\mathrm{ea},\mathrm{af}\}$
\State \Return $R_{\mathrm{CAVE}}(\tau)=\lambda_{\mathrm{bu}}C_{\mathrm{bu}}+\lambda_{\mathrm{ea}}C_{\mathrm{ea}}+\lambda_{\mathrm{af}}C_{\mathrm{af}}$
\end{algorithmic}
\end{algorithm}

As shown in Algorithm~\ref{alg:cave_credit}, for each type of process credit, we first aggregate step-level credits into trajectory-level credits. Let $m\in\{\mathrm{bu},\mathrm{ea},\mathrm{af}\}$, and let $c_t^m$ denote the local credit at step $t$. The \modelname process reward is defined as

\vspace{-15pt}
\begin{equation*}
\begin{aligned}
C_m(\tau)=\sum_{t=1}^{T_\tau} w_t^m\,\phi_m(c_t^m),
\qquad
R_{\mathrm{CAVE}}(\tau)
=
\lambda_{\mathrm{bu}} C_{\mathrm{bu}}(\tau)
+
\lambda_{\mathrm{ea}} C_{\mathrm{ea}}(\tau)
+
\lambda_{\mathrm{af}} C_{\mathrm{af}}(\tau),
\end{aligned}
\end{equation*}

\vspace{-10pt}

where $w_t^m$ denotes a round-dependent weight used to encourage earlier evidence progression, and $\phi_m(\cdot)$ denotes a bounded transformation that clips extreme local credits to stabilize GRPO training. Given an input $x=(I,q)$, we sample a group of rollouts $\{\tau_i\}_{i=1}^{G}$ from the old policy $\pi_{\theta_{\mathrm{old}}}$, and compute the group-relative advantage based on $R_{\mathrm{CAVE}}$:

\vspace{-13pt}
\begin{equation*}
\hat A_i=
\frac{
R_{\mathrm{CAVE}}(\tau_i)-\mathrm{mean}_{j=1}^{G} R_{\mathrm{CAVE}}(\tau_j)
}{
\mathrm{std}_{j=1}^{G} R_{\mathrm{CAVE}}(\tau_j)+\delta
}.
\end{equation*}

\vspace{-8pt}

We then update the policy using the GRPO objective \citep{shao2024deepseekmath}. Optimization is applied only to the assistant-generated tokens; the observations returned by tools, system prompts, and environment-injected information serve only as contextual conditions, with their response masks set to $0$, and do not participate in policy-gradient computation. In practice, we add lightweight task-anchoring terms, such as answer correctness, format validity, and round-number constraints, to maintain training stability. In this way, \modelname extends sparse supervision centered on terminal correctness into an optimization signal for visual evidence progression under group-relative comparison.

\vspace{-10pt}
\section{TRACER-Bench}
\label{sec:tracer_bench}
\vspace{-10pt}

\begin{figure*}[t]
\centering

\begin{minipage}[t]{0.28\textwidth}
\vspace{0pt}
\centering
\includegraphics[width=\linewidth]{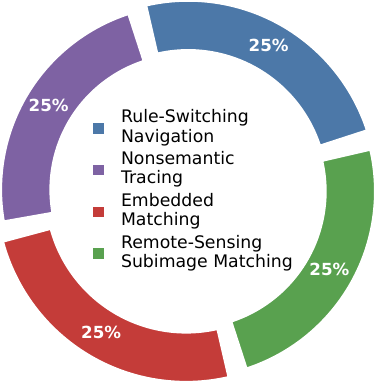}
\vspace{-5mm}

\captionof{figure}{Scenario distribution of TRACER-Bench.}
\label{fig:tracer_bench_distribution}
\end{minipage}
\hfill
\begin{minipage}[t]{0.68\textwidth}
\vspace{0pt}
\centering
\small
\setlength{\tabcolsep}{3pt}

\captionof{table}{
Comparison between TRACER-Bench and related vision-centric multimodal benchmarks.
}
\label{tab:benchmark_comparison}

\begin{tabularx}{\linewidth}{@{}
>{\raggedright\arraybackslash}X
>{\centering\arraybackslash}p{0.20\linewidth}
>{\centering\arraybackslash}p{0.14\linewidth}
>{\centering\arraybackslash}p{0.24\linewidth}
@{}}
\toprule
Benchmark 
& \makecell{Cross-region\\dependency}
& \makecell{Difficulty\\control}
& \makecell{Explicit process\\evidence} \\
\midrule
Tunnel Vision~\citep{berman2025tunnelvision} 
& $\checkmark$ 
& $\checkmark$ 
& $\times$ \\
BabyVision~\citep{chen2026babyvision} 
& Partial 
& Partial 
& $\times$ \\
BLINK~\citep{fu2024blink} 
& Partial 
& Partial 
& $\times$ \\
HallusionBench~\citep{guan2024hallusionbench} 
& Partial 
& $\times$ 
& $\times$ \\
RealWorldQA~\citep{xai2024realworldqa} 
& Partial 
& $\times$ 
& $\times$ \\
MMStar~\citep{chen2024mmstar} 
& Partial 
& Partial 
& $\times$ \\
\midrule
TRACER-Bench 
& $\checkmark$ 
& $\checkmark$ 
& $\checkmark$ \\
\bottomrule
\end{tabularx}
\end{minipage}

\vspace{5pt}
\end{figure*}

\noindent\textbf{Scenario coverage and construction.} Existing benchmarks with nonlocal visual nature such as Tunnel Vision and BabyVision provide important references for studying the basic visual capabilities of VLMs \citep{berman2025tunnelvision,chen2026babyvision}. However, these benchmarks are difficult to simultaneously satisfy the requirements of scenario coverage, difficulty control, and process-evidence analysis. To this end, we construct {TRACER-Bench} (Text-Resistant Cross-Region Evidence Reasoning), a compact and controllable fragmented visual reasoning benchmark containing 980 VQA samples. It covers common cross-region dependency patterns with spatially fragmented and semantically weakly separable evidence, including three abstract scenarios and one real-world scenario: {Rule-Switching Navigation} (VJump), {Nonsemantic Tracing} (LT), {Embedded Matching} (Match), and {Remote-Sensing Subimage Matching} (MME-RS), corresponding to new dimensions of rule-switching dependency chains, non-semantic path tracing, local--global structural matching, and real-world remote-sensing subimage matching. In addition, TRACER-Bench annotates task-critical visual evidence in textual form, supporting in-depth offline process analysis of the reasoning process, while not being provided as model input. Scenario distribution and benchmark comparisons are shown in Figure~\ref{fig:tracer_bench_distribution} and Table~\ref{tab:benchmark_comparison}. Scenario design, dataset statistics, quality control are provided in Appendix~\ref{app:tracer_bench_details}.

\noindent\textbf{Training corpus and data isolation.} We further construct a training corpus for CAVE reinforcement learning and use \texttt{perception} to compute the evidence acquisition credit. The training data shares the same four scenario families with TRACER-Bench, but is completely isolated from the test set in terms of random seeds, layouts, image regions, and concrete instances, in order to avoid instance leakage and evaluate generalization to unseen layouts and dependency structures. To improve structural coverage, we additionally construct non-overlapping Tunnel-Vision-style synthetic data \citep{berman2025tunnelvision}, and supplement the \texttt{perception} field to unify it into the \texttt{prompt-answer-perception} format. This data is used only as an auxiliary training corpus, is not part of TRACER-Bench, and does not overlap with the public Tunnel Vision evaluation set. Detailed construction and split protocols are provided in Appendix~\ref{app:training_data_construction}.

\vspace{-10pt}
\section{Experiments}
\label{sec:experiments}
\vspace{-8pt}
\subsection{Experimental setup}
\vspace{-5pt}

\textbf{Baselines.}
We compare six categories of methods: Human and Random as the reference upper bound and chance-level reference, respectively; GPT-5.2 and Gemini 3.1 Pro as strong closed-source model references \citep{openai2025gpt52,google2026gemini31pro}; Qwen2.5-VL-7B, Qwen3-VL-8B, InternVL3.5-8B, and MiniCPM-V 4.5 as open-source general-purpose VLMs \citep{bai2025qwen25vl,bai2025qwen3vl,wang2025internvl35,yu2025minicpmv45}; R1-OneVision-7B, ThinkLite-VL, and VLAA-Thinker as recent reasoning-oriented VLMs~\citep{yang2025r1onevision,wang2025thinklite,chen2025sftorrl}; DeepEyes as a tool-assisted look-again method~\citep{zheng2025deepeyes}; and Belief-Only as a same-backbone process-reward model for examining whether single answer-side information gain is sufficient. CAVE-3B and CAVE-7B are used to evaluate parameter efficiency and the full method performance, respectively.




\textbf{Benchmarks.}
We evaluate model capabilities at two levels: fragmented visual evidence reasoning and general multimodal benchmarks. The former includes Tunnel Vision, BabyVision, and our constructed TRACER-Bench, containing 1,350, 388, and 980 samples, respectively \citep{berman2025tunnelvision,chen2026babyvision}. TRACER-Bench consists of TRACER-Core, which contains three abstract and controllable scenarios, and MME-RS, which targets real-world remote-sensing scenarios. The latter includes MMStar, RealWorldQA, HallusionBench-Image, BLINK, MathVista-mini, and MathVerse-mini, which are used to evaluate out-of-domain transfer and the preservation of general capabilities \citep{chen2024mmstar,xai2024realworldqa,guan2024hallusionbench,fu2024blink,lu2024mathvista,zhang2024mathverse}. Unless otherwise specified, all methods adopt the same input format, answer extraction, and evaluation protocol.

\textbf{Training details.}
CAVE uses Qwen2.5-VL as the base model and follows a two-stage training procedure on 8 A800 GPUs: cold-start training on 10k DeepEyes-style samples, followed by process-reward training on 5,000 TRACER-style samples with process evidence annotations \citep{bai2025qwen25vl,zheng2025deepeyes,shao2024deepseekmath}. Detailed training and reward configurations are provided in Appendix~\ref{app:training_details}.


\begin{table*}[t]
\centering
\setlength{\tabcolsep}{3.2pt}
\renewcommand{\arraystretch}{1.12}
\caption{Main results on fragmented vision reasoning benchmarks. Results are grouped by Tunnel Vision, BabyVision, and TRACER-Bench. Models marked with \textsuperscript{*} are Instruct versions.}
\label{tab:main_baselines}
\resizebox{\textwidth}{!}{
\begin{tabular}{@{}ll|ccc|cccc|cccc|c@{}}
\toprule
\multirow{2}{*}{Model}
& \multirow{2}{*}{Param}
& \multicolumn{3}{c}{Tunnel Vision}
& \multicolumn{4}{c}{BabyVision}
& \multicolumn{4}{c}{TRACER-Bench}
& \multirow{2}{*}{\makecell{Overall\\Avg.}} \\
\cmidrule(lr){3-5} \cmidrule(lr){6-9} \cmidrule(lr){10-13}
& 
& ObjReID & VS & Circuit
& FD & SP & VPR & VT
& {LT} & Match & {VJump} & {MME-RS}
& \\
\midrule

Human & -- 
& 100 & 100 & 99.0 
& 94.1 & 94.1 & 94.1 & 94.1 
& 100 & 97.6 & 95.2 & 99.0 & 98.3 \\
Random & -- 
& 50.0 & 12.5 & 4.00 
& 10.3 & 10.3 & 10.3 & 10.3 
& 16.2 & 50.0 & 12.1 & 50.0 & 24.1 \\
\midrule

GPT-5.2 & -- 
& 71.8 & 68.4 & 52.9 
& 27.3 & 35.2 & 54.9 & 34.9
& 58.4 & 68.2 & 52.2 & 72.7 & 61.8 \\
Gemini 3.1 Pro & -- 
& 67.3 & 72.0 & 46.4 
& 46.2 & 53.7 & 53.9 & 43.4
& 53.5 & 64.5 & 44.1 & 67.8 & 56.4 \\
\midrule

Qwen2.5-VL\textsuperscript{*} & 7B
& 44.4 & 11.1 & 17.6 
& 10.4 & 13.1 & 13.7 & 7.79
& 22.9 & 53.5 & 15.9 & 43.7 & 25.9 \\
Qwen3-VL\textsuperscript{*} & 8B 
& 48.2 & {13.6} & 22.4 
& 13.5 & 14.3 & 11.7 & 9.09
& \textbf{35.9} & 55.1 & 15.1 & 52.7 & 30.1 \\
InternVL3.5\textsuperscript{*} & 8B 
& 45.1 & \underline{15.3} & 20.4 
& \underline{14.1} & \textbf{21.9} & 19.6 & 10.4
& 26.1 & 41.2 & \textbf{23.7} & \underline{58.4} & 29.2 \\
MiniCPM-V 4.5 & 8B
& 50.0 & 11.8 & 17.8 
& 12.3 & 19.8 & 19.6 & 9.09
& 30.2 & 52.7 & 16.7 & 55.1 & 29.2 \\
\midrule

R1-OneVision & 7B
& \underline{52.6} & 8.00 & 16.7 
& 9.20 & 16.5 & \underline{23.5} & \underline{11.7}
& 22.0 & 43.7 & 7.80 & 51.8 & 26.0 \\
VLAA-Thinker & 7B 
& 46.9 & 10.9 & 22.2 
& 12.3 & \underline{20.9} & 19.6 & \underline{11.7}
& 22.0 & 43.7 & 14.3 & 47.8 & 27.0 \\
ThinkLite-VL & 7B 
& 46.9 & 10.2 & \underline{22.7} 
& 11.0 & 17.6 & 21.6 & \underline{11.7}
& 22.0 & 45.3 & 17.6 & 50.2 & 27.4 \\
\midrule

DeepEyes & 7B 
& 49.3 & 13.3 & 19.1 
& 11.5 & 18.7 & 22.1 & 10.9
& 23.7 & 49.4 & 15.9 & 54.3 & 28.5 \\
\midrule
\rowcolor{blue!5}
CAVE Belief-Only & 7B
& 49.7 & 12.2 & 20.9 
& 12.9 & 18.7 & 15.7 & 9.10
& 28.8 & 52.8 & 16.8 & 55.2 & 29.6 \\
\rowcolor{blue!5}
CAVE-3B & 3B
& 50.2 & 12.9 & 21.1 
& 13.5 & 17.6 & \underline{23.5} & 10.4
& 27.2 & \underline{56.0} & 18.4 & 57.6 & \underline{30.5} \\
\rowcolor{blue!5}
CAVE-7B & 7B 
& \textbf{53.5} & \textbf{16.7} & \textbf{23.8} 
& \textbf{15.3} & 18.7 & \textbf{25.5} & \textbf{14.3}
& \underline{31.4} & \textbf{58.4} & \underline{20.8} & \textbf{59.2} & \textbf{33.3} \\
\rowcolor{blue!5}
\(\Delta\) (vs Qwen2.5-VL) & 7B 
& +9.1 & +5.6 & +6.2 
& +4.9 & +5.6 & +11.8 & +6.5
& +8.5 & +4.9 & +4.9 & +15.5 & +7.4 \\
\bottomrule
\end{tabular}
}
\vspace{-15pt}
\end{table*}

\vspace{-10pt}
\subsection{Main results}
\vspace{-8pt}

We first compare CAVE with different types of baselines on reasoning tasks with fragmented vision, with results shown in Table~\ref{tab:main_baselines}. CAVE achieves the best overall average performance among comparable open-source baselines and significantly outperforms Qwen2.5-VL-7B with the same backbone \citep{bai2025qwen25vl}, indicating that its gains mainly come from structured optimization of the intermediate visual evidence progression process, rather than a stronger pretrained model. Compared with the tool-assisted DeepEyes~\citep{zheng2025deepeyes} and reasoning-oriented VLMs such as ThinkLite-VL and VLAA-Thinker~\citep{wang2025thinklite,chen2025sftorrl}, CAVE achieves stronger overall performance, suggesting that neither tool-based re-observation nor generic reasoning-oriented training is sufficient for robust evidence acquisition, state maintenance, and cross-region integration.

Furthermore, Belief-Only outperforms most base and tool-augmented models \citep{bai2025qwen3vl,wang2025internvl35,yang2025r1onevision,zheng2025deepeyes,wang2025thinklite,chen2025sftorrl}, but remains clearly weaker than the full CAVE, indicating that belief gain is an effective process signal, yet is insufficient by itself to cover precursor evidence acquisition and visual-action focusing. CAVE-3B exceeds multiple 7B/8B open-source baselines in overall average performance \citep{bai2025qwen3vl,wang2025internvl35}, showing that structured process-credit training can still induce effective visual evidence progression behaviors with a smaller-parameter backbone. Together, these results demonstrate that the advantage of CAVE does not come from a single information-gain reward, model scale, or a specific tool format, but from structured modeling of the visual evidence progression process.

We further examine the out-of-domain transfer and capability preservation of CAVE on general multimodal evaluations, with results shown in Table~\ref{tab:generalization_results} \citep{chen2024mmstar,xai2024realworldqa,guan2024hallusionbench,fu2024blink,lu2024mathvista,zhang2024mathverse}. CAVE achieves the best results on four of the six general benchmarks and the second-best result on two benchmarks, indicating that process optimization centered on visual evidence exploration does not weaken general multimodal capabilities and brings positive transfer to several vision-language reasoning scenarios.

In summary, CAVE not only brings stable same-backbone improvements on fragmented vision reasoning tasks, but also demonstrates strong transfer capability on general multimodal evaluations, indicating that what it optimizes is not a specific benchmark or data template, but a more transferable capability of visual evidence exploration.

\begin{table*}[t]
\vspace{10pt}
\centering
\small
\setlength{\tabcolsep}{5pt}
\renewcommand{\arraystretch}{1.1}
\caption{Comparison of model performance on general multimodal benchmarks. HallusionBench-Image (avg.) reports the average score over the official HallusionBench metrics.}
\label{tab:generalization_results}
\resizebox{\textwidth}{!}{
\begin{tabular}{lcccccc}
\toprule
\multirow{2}{*}{Model} & \multicolumn{6}{c}{Generalization Performance} \\
\cmidrule(lr){2-7}
& MMStar & RealWorldQA & Hallu.-Img (avg.) & BLINK & MathVista (mini) & MathVerse (mini) \\
\midrule
Qwen2.5-VL (7B)      & 61.5 & 59.4 & 60.3 & 58.5 & 66.7 & 48.1  \\
Qwen3-VL-8B-Instruct & 64.3 & \underline{69.3} & 62.5 & \textbf{63.3} & \underline{70.6} & \textbf{60.9}  \\
InternVL3.5-8B       & \underline{64.5} & 61.6 & 57.9 & 59.8 & 69.6 & 53.2  \\
DeepEyes-7B          & 63.6 & 68.4 & \underline{63.1} & 56.4 & 70.3 & 49.2  \\
\midrule
CAVE-7B              & \textbf{65.2} & \textbf{69.7} & \textbf{64.1} & \underline{61.5} & \textbf{71.2} & \underline{55.7}  \\
\(\Delta\) (vs. Qwen2.5-VL-7B) 
                     & +3.7 & +10.3 & +3.8 & +3.0 & +4.5 & +7.6 \\
\bottomrule
\end{tabular}
}
\vspace{2mm}
\begin{minipage}{0.98\textwidth}
\footnotesize
\end{minipage}
\vspace{-3pt}
\end{table*}

\vspace{-10pt}
\subsection{Ablation studies}
\vspace{-5pt}

\begin{table}[t]
\centering
\small
\setlength{\tabcolsep}{3.5pt}
\renewcommand{\arraystretch}{1.08}
\caption{Ablation studies on structured credits and training sources, evaluating credit complementarity and the effect of data composition on fragmented vision reasoning.}
\label{tab:ablation_main}

\begin{subtable}[t]{0.465\linewidth}
\centering
\caption{Structured credits ablation.}
\label{tab:reward_ablation}
\begin{tabular*}{\linewidth}{@{\extracolsep{\fill}}lcccc@{}}
\toprule
Method 
& \makecell[c]{Tunnel\\Vision} 
& \makecell[c]{Baby\\Vision}
& \makecell[c]{TRACER\\-Core}
& \makecell[c]{MME\\-RS} \\
\midrule
Cold-Start        & 25.2 & 11.6 & 30.3 & 46.1\\
Correct Only      & 26.7 & 11.3 & 31.2 & 49.4\\
w/o Belief        & 28.9 & 15.6 & 34.1 & 56.7\\
w/o Evidence      & 29.6 & 15.3 & 34.7 & 55.9\\
w/o Focusing      & 29.3 & 16.5 & 35.2 & 57.6\\
Full CAVE         & 31.3 & 17.3 & 36.9 & 59.2\\
\bottomrule
\end{tabular*}
\end{subtable}
\hfill
\begin{subtable}[t]{0.515\linewidth}
\centering
\caption{Training-source ablation.}
\label{tab:dataset_ablation}
\setlength{\tabcolsep}{3pt}
\begin{tabular*}{\linewidth}{@{\extracolsep{\fill}}lcccc@{}}
\toprule
Training Data
& \makecell[c]{Tunnel\\Vision}
& \makecell[c]{Baby\\Vision}
& \makecell[c]{TRACER\\-Core}
& \makecell[c]{MME\\-RS} \\
\midrule
Cold-start        & 25.2 & 11.6 & 30.3 & 46.1\\
DeepEyes data     & 29.3 & 15.1 & 33.9 & 53.5\\
w/o TV            & 29.9 & 16.6 & 34.0 & 51.8\\
w/o RS            & 30.6 & 16.4 & 35.5 & 56.7\\
w/o TRACER-Core   & 30.3 & 15.7 & 34.8 & 57.6\\
Full Source       & 31.3 & 17.3 & 36.9 & 59.2\\
\bottomrule
\end{tabular*}
\end{subtable}

\vspace{-10pt}
\end{table}

We conduct ablation analyses from two perspectives: structured credits design and training-source composition. All variants use the same base model, interleaved reasoning structure, and training configuration, changing only the process credit terms or the data sources used in the second-stage training.

\textbf{Structured credits ablation.}
Table~\ref{tab:reward_ablation} evaluates the contribution of structured process credits. {Cold-start} learns the basic multi-round tool-use format and does not undergo CAVE reinforcement learning; {Correct Only} uses the final-answer correctness reward; and the remaining variants remove belief update, evidence acquisition, or adaptive focus control, respectively. The results show that both Cold-start and Correct Only are clearly weaker than the full CAVE, indicating that the gains do not merely come from interleaved tool use or terminal supervision. Removing any process credit also leads to performance degradation, demonstrating the complementarity of the three signals. Among them, w/o Belief usually causes the most pronounced degradation, suggesting that answer-belief update is the most stable progress signal; evidence acquisition and focusing also bring consistent gains. Overall, the full CAVE achieves the best or most balanced performance, verifying the necessity of structured credit assignment for visual evidence exploration.

\textbf{Training-source ablation.}
Table~\ref{tab:dataset_ablation} analyzes the role of training sources. We compare training configurations that use DeepEyes general-reasoning data supplemented with perception fields, as well as those that remove TV (Tunnel Vision), TRACER-Core, or MME-RS, respectively. Even when a certain source is removed, the model still improves over cold-start on the corresponding test domain, alleviating the concern that CAVE merely memorizes a specific generator or data-source template. Meanwhile, the full mixed-data setting achieves the best or most balanced performance, indicating that different sources provide complementary coverage: Tunnel-Vision-style data provides non-local reasoning forms \citep{berman2025tunnelvision}, TRACER-Core covers abstract and controllable dependencies, and Remote-Sensing introduces scale, rotation, and complex background variations in real images. Together, the two groups of ablations show that the core gains of CAVE come from structured process credits, while multi-source training further improves coverage and robustness.

\vspace{-10pt}
\subsection{Process-credit validation and difficulty robustness}
\vspace{-8pt}

\begin{figure*}[t]
    \centering
    \begin{subfigure}[t]{0.24\textwidth}
        \centering
        \caption{Credit scatter}
        \includegraphics[width=\linewidth]{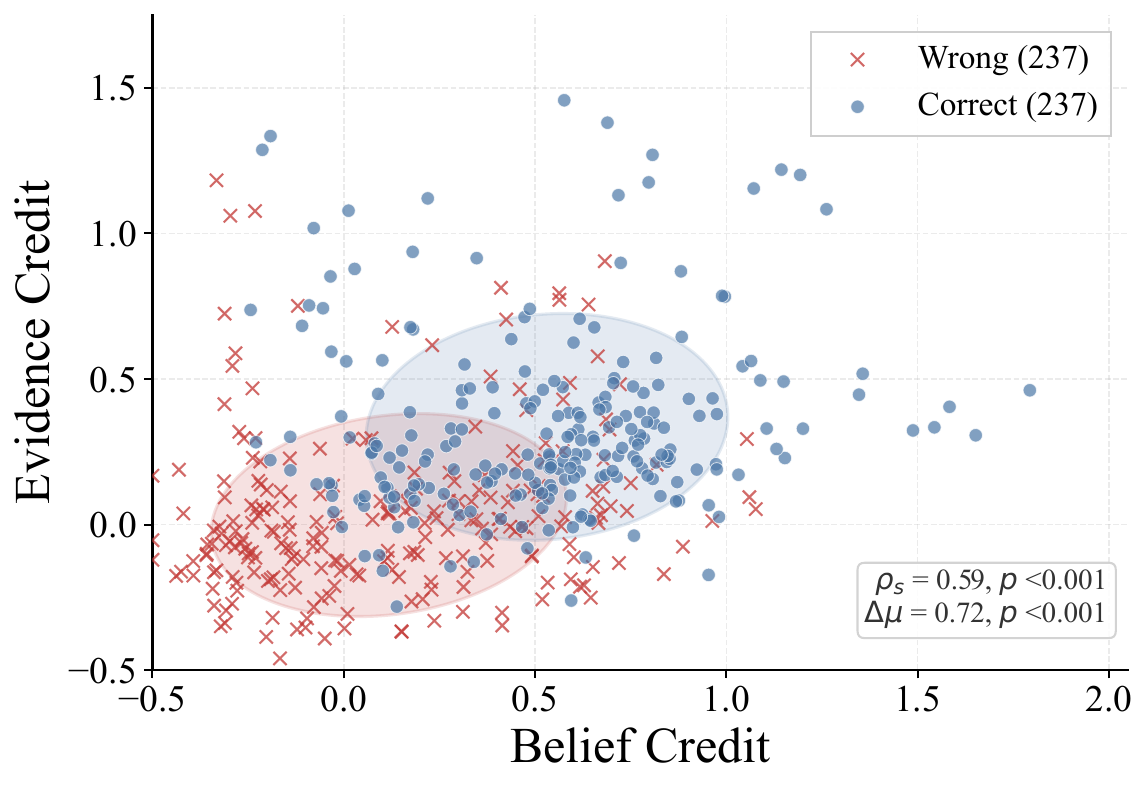}
        \label{fig:balanced_reward_scatter}
    \end{subfigure}
    \hfill
    \begin{subfigure}[t]{0.24\textwidth}
        \centering
        \caption{Credit quantiles}
        \includegraphics[width=\linewidth]{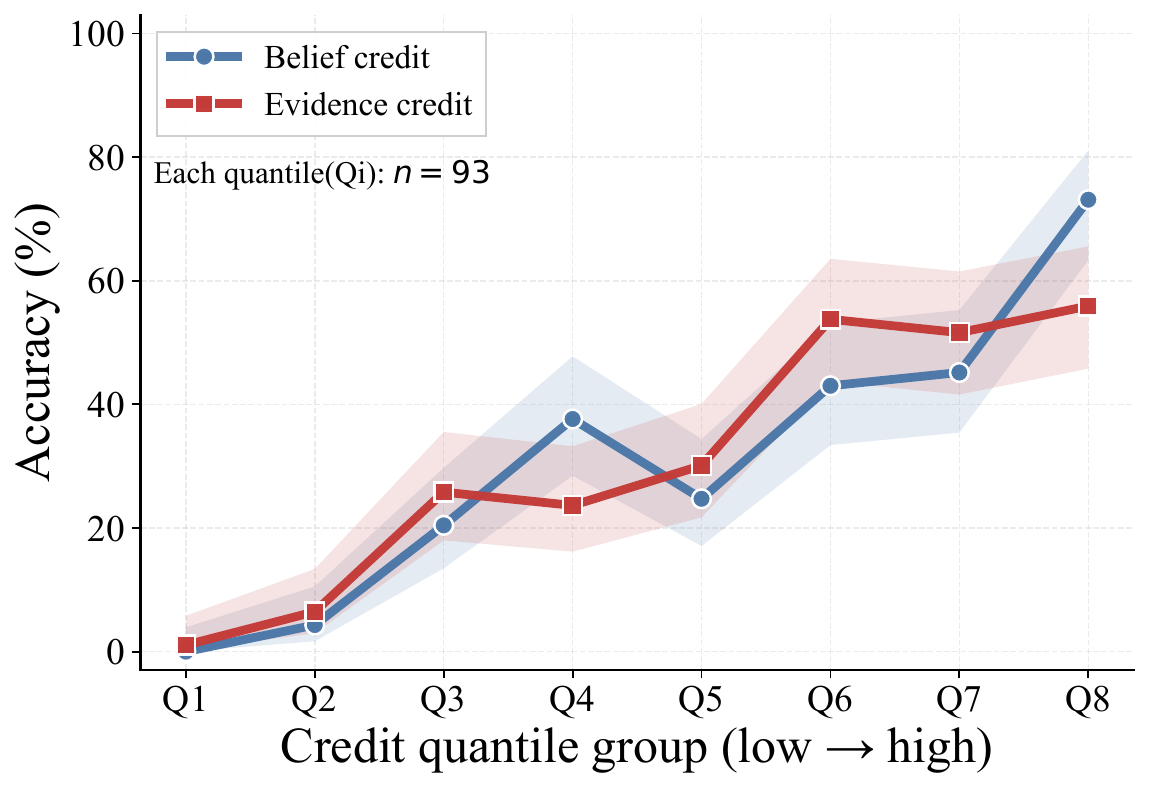}
        \label{fig:gain_quantile_accuracy}
    \end{subfigure}
    \hfill
    \begin{subfigure}[t]{0.24\textwidth}
        \centering
        \caption{Reasoning depth}
        \includegraphics[width=\linewidth]{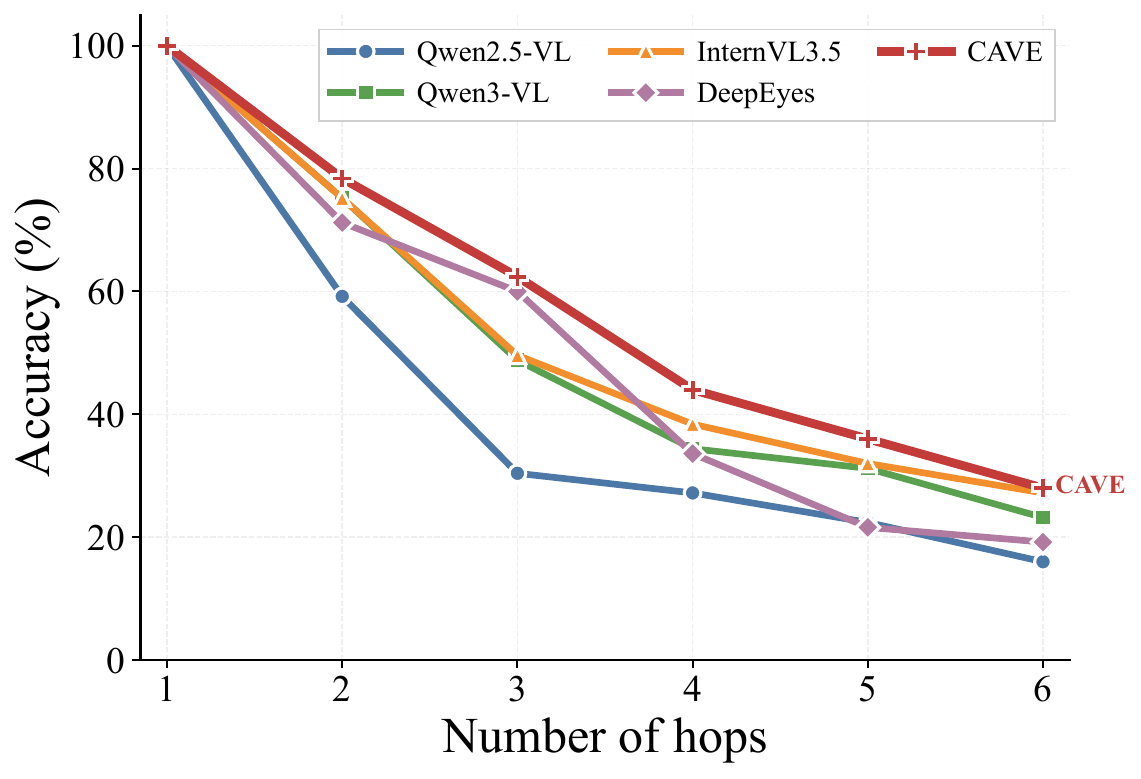}
        \label{fig:jump_depth_accuracy}
    \end{subfigure}
    \hfill
    \begin{subfigure}[t]{0.24\textwidth}
        \centering
        \caption{Region distance}
        \includegraphics[width=\linewidth]{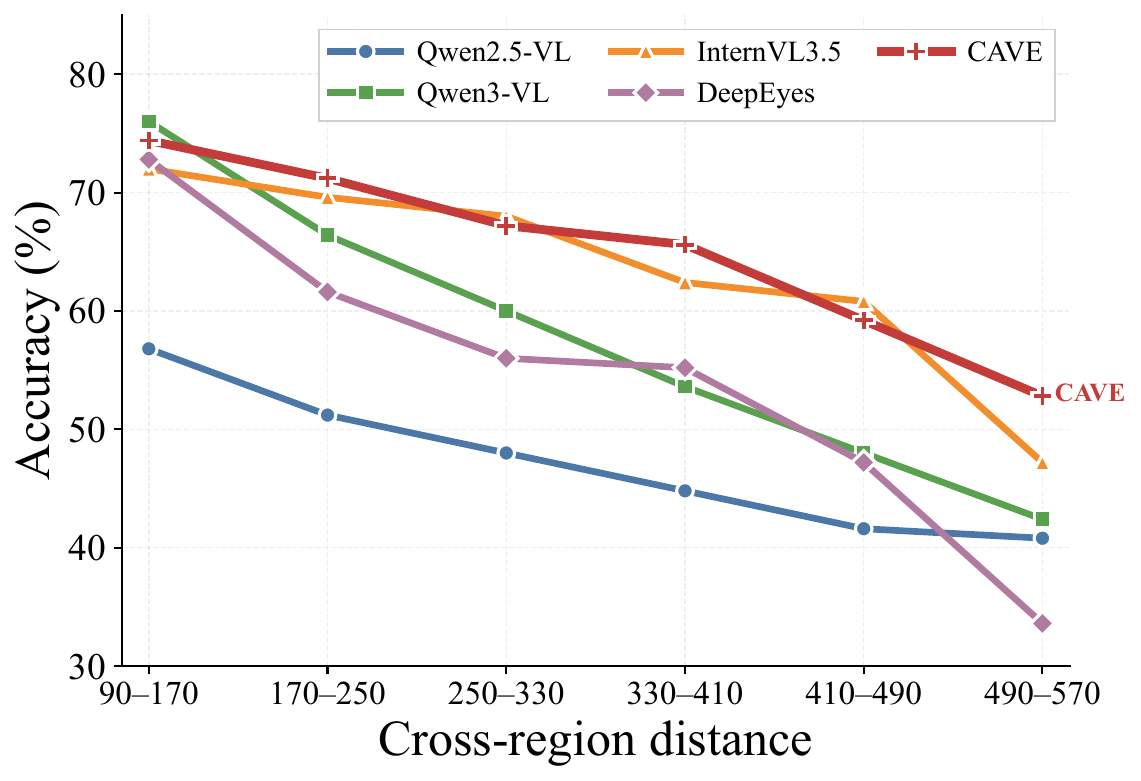}
        \label{fig:distance_accuracy}
    \end{subfigure}
\vspace{-10pt}
\caption{
Process-credit validation and difficulty analysis. 
(a) Distribution of correct and wrong trajectories in the belief-credit and evidence-credit space, with $\rho_s$ and $\Delta\mu$ reporting the credit--correctness correlation and mean credit gap.
(b) Accuracy after grouping samples into eight equal-size quantiles from low to high credit values, with shaded Wilson confidence intervals.
(c,d) Accuracy on 1,500 fragmented-vision samples grouped by required reasoning hops and cross-region distance, respectively, where CAVE shows more stable performance under increasing difficulty.
}
    \label{fig:process_reward_and_difficulty}
\vspace{8pt}
\end{figure*}

Beyond the main results and ablation studies, we further validate the consistency between process credits and successful reasoning, and analyze the robustness of CAVE under different fragmented vision reasoning difficulty conditions.

\textbf{Validity of process gains.}
To avoid self-validation bias on the model trained by CAVE, we conduct the analysis using an interleaved reasoning model that has only undergone cold-start training, which has not yet been directly optimized by belief gain or evidence gain \citep{zheng2025deepeyes,wang2025visualprm,wang2026igpo}. As shown in Figure~\ref{fig:process_reward_and_difficulty}(a)(b), correct trajectories are overall more concentrated in regions with higher belief/evidence gain; after grouping by gain percentiles, the final accuracy also generally increases as the gain becomes larger. Among them, belief gain exhibits a more stable positive correlation, indicating that it can characterize whether the answer formation process progresses in the correct direction; evidence gain shows certain fluctuations, but is also positively correlated with correctness, suggesting that the acquisition of key intermediate evidence contributes to final reasoning. These results indicate that the two types of state-level process signals are not arbitrary heuristics, but are empirically consistent with successful visual evidence reasoning progression.

\textbf{Robustness under increasing difficulty.}
We further analyze task difficulty from two dimensions: cross-region distance and reasoning hops, which characterize the spatial span of key evidence and the depth of the dependency chain, respectively \citep{berman2025tunnelvision,chen2026babyvision}. As shown in Figure~\ref{fig:process_reward_and_difficulty}(c)(d), as distance and the number of hops increase, the accuracy of all models generally decreases, indicating that long-range evidence localization, multi-step state maintenance, and region transition substantially increase reasoning difficulty. In contrast, CAVE maintains leading performance in most intervals and exhibits slower performance degradation, suggesting that its benefits are not limited to simple samples, but can still advance the visual evidence chain more stably under longer-range and deeper cross-region dependencies. These results further indicate that the advantages of CAVE mainly come from more effective evidence selection and state maintenance, rather than improvements only on low-difficulty samples.

\vspace{-10pt}
\subsection{Process behavior, efficiency, and training stability}
\vspace{-8pt}

\begin{table}[t]
\centering
\small
\setlength{\tabcolsep}{3.8pt}
\renewcommand{\arraystretch}{1.15}

\newcommand{\budgettablewidth}{0.98\linewidth}

\caption{Budget-controlled comparison of process behavior on fragmented vision reasoning benchmarks. Avg. tokens counts only assistant-generated tokens, excluding tool observations and environment-injected messages. Useful zoom rate is used as a process diagnostic and denotes the proportion of zoom actions that yield positive belief or evidence gain after observation.}
\label{tab:budget_controlled}

\begin{tabular*}{\budgettablewidth}{@{\extracolsep{\fill}}lcccccc@{}}
\toprule
Method
& \makecell[c]{Acc.\\(\%)}
& \makecell[c]{Avg.\\Tokens}
& \makecell[c]{Avg.\\Rounds}
& \makecell[c]{Avg.\\Tool Calls}
& \makecell[c]{Useful\\Zoom Rate}
& \makecell[c]{Time / Sample\\(s)}
\\
\midrule
Cold-start & 26.5 & 1935 & 3.12 & 2.46 & 21.4 & 5.24 \\
DeepEyes   & 28.5 & 1643 & \textbf{2.48} & \textbf{1.97} & 43.1 & 4.53 \\
CAVE-7B       & \textbf{33.3} & \textbf{1566} & 2.51 & 2.03 & \textbf{55.4} & \textbf{4.27} \\
\bottomrule
\end{tabular*}

\vspace{5pt}
\end{table}

We further conduct diagnostic analyses to verify whether the gains of CAVE come from more reasonable process behavior, rather than longer outputs, more tool calls, or unstable training. First, the reasoning trajectory visualizations in Appendix~\ref{sec:case_study_rollout} show that, compared with the cold-start model and general look-again methods \citep{zheng2025deepeyes,zhang2025chainoffocus}, CAVE is more likely to localize task-relevant regions and continuously update intermediate states around the acquired evidence, exhibiting clearer evidence-grounded reasoning behavior. Second, we report the average number of generated tokens, reasoning rounds, zoom calls, useful zoom rate, and inference time in Table~\ref{tab:budget_controlled}. The results show that CAVE achieves higher accuracy under a comparable or even lower token/tool budget than other interleaved reasoning or tool-augmented methods \citep{su2025openthinkimg,zheng2025deepeyes,zhang2025chainoffocus}, and obtains a higher useful zoom rate, indicating that its improvement mainly comes from more effective visual focusing and evidence acquisition. Finally, the training dynamics show that the structured rewards and their components improve steadily, as shown in Appendix~\ref{app:budget_controlled}, while the KL remains within a controllable range, with no obvious policy collapse or abnormal drift observed \citep{shao2024deepseekmath,sheng2024hybridflow}. Taken together, the improvements of CAVE are reflected not only in final accuracy, but also in more effective evidence localization, higher-quality tool use, and a stable reinforcement learning optimization process.

\vspace{-15pt}
\section{Conclusion}
\label{sec:conclusion}
\vspace{-10pt}

This paper studies Fragmented Visual Reasoning, where VLMs must integrate spatially distributed and semantically weakly separable evidence under low language priors. To systematically evaluate this challenge, we construct \datasetname with diverse cross-region scenarios and verifiable intermediate evidence. To address the difficulty of attributing interleaved reasoning trajectories, we propose \modelname, which assigns process credits for visual evidence exploration from three action-level dimensions: belief update, evidence acquisition, and adaptive focus control. These credits guide the model to discover task-critical evidence, update reasoning states, and focus on informative regions. Experiments show that \modelname consistently improves fragmented visual reasoning performance while maintaining strong general multimodal capabilities. Further analyses show that \modelname advances visual evidence chains more effectively and exhibits stronger robustness under long-range and deep cross-region dependencies.

% ...

\bibliographystyle{unsrtnat}
\bibliography{references}

\appendix
\newpage
\section{Case study on rollout trajectories}
\label{sec:case_study_rollout}

\begin{figure}[!htbp]
    \centering
    \includegraphics[width=\linewidth]{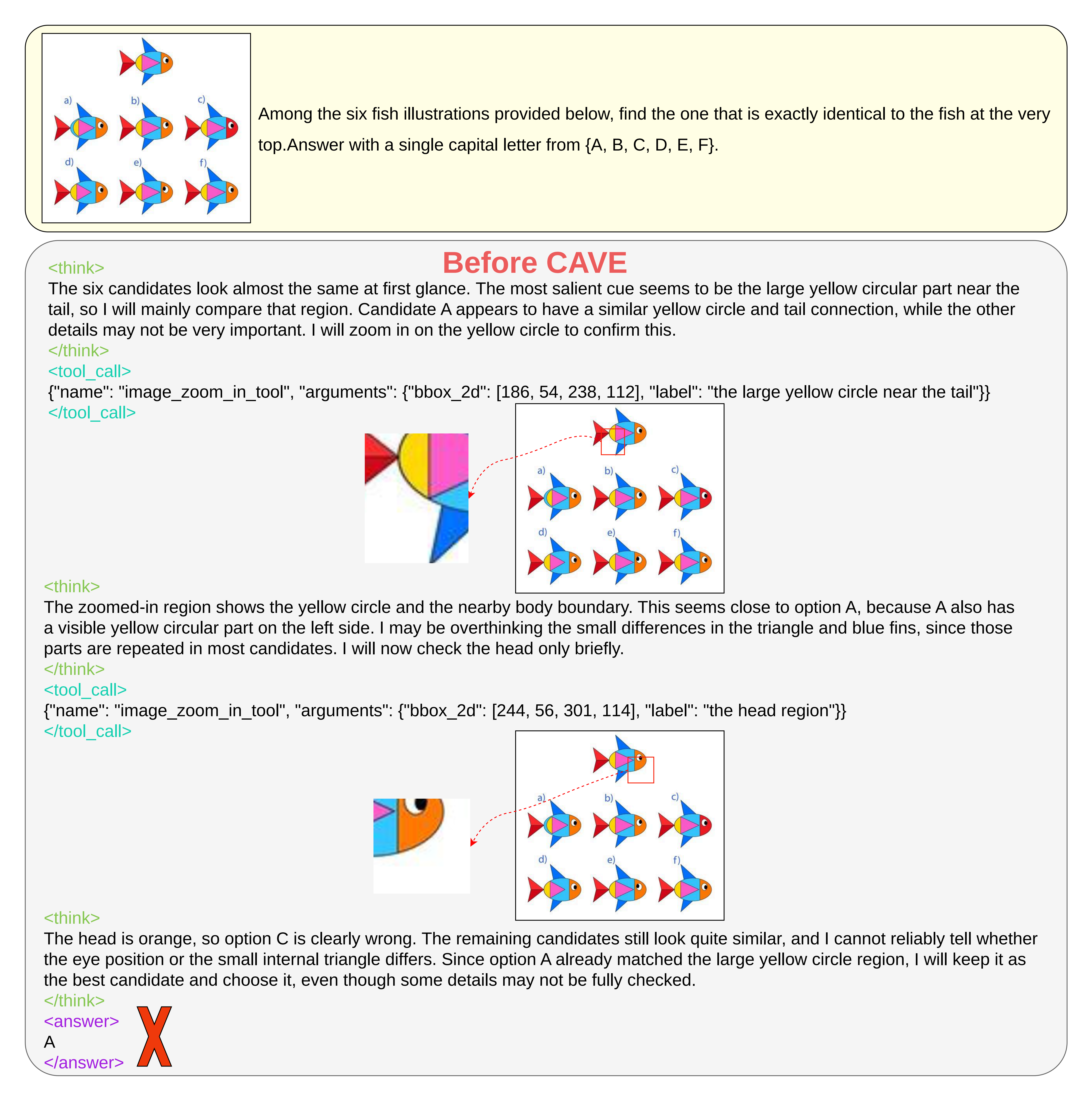}
    \caption{
    Case study of rollout trajectories before CAVE training.
    }
    \label{fig:case_study_rollout_before}
\end{figure}

\begin{figure}[!htbp]
    \centering
    \includegraphics[width=\linewidth]{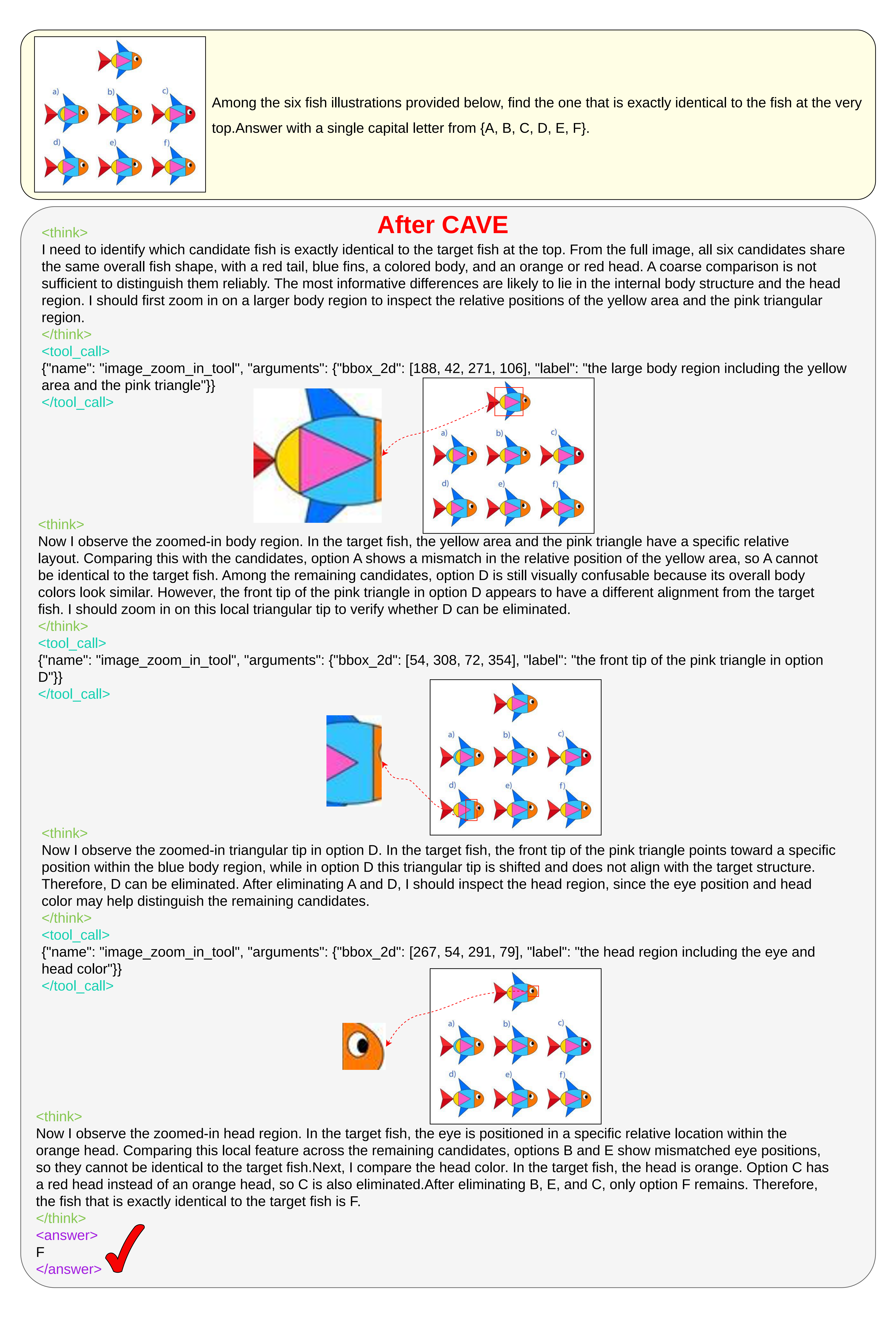}
    \caption{
    Case study of rollout trajectories after CAVE training.
    }
    \label{fig:case_study_rollout_after}
\end{figure}

Figure~\ref{fig:case_study_rollout_before} and Figure~\ref{fig:case_study_rollout_after} present a comparison of rollout trajectories before and after CAVE training. The example requires the model to identify which candidate fish is exactly identical to the target fish. The correct decision depends on several confusable local cues. We use this case to visualize how different models acquire visual evidence through zoom-in actions and how such evidence affects subsequent elimination of candidates.

\section{TRACER-Bench}
\label{app:tracer_bench_details}

This section provides additional details of TRACER-Bench, including scenario design, dataset statistics and quality control. TRACER-Bench is designed to evaluate cross-region dependency reasoning under controlled yet diverse visual structures. Each sample follows a unified \texttt{prompt-answer-perception} format, where the \texttt{perception} field records key intermediate visual evidence required to solve the task. This design allows the benchmark to support both final-answer evaluation and process-level analysis.

\subsection{Scenario design}
\label{app:tracer_bench_design}
 
Table~\ref{tab:tracer_bench_scenarios} summarizes four scenario families in TRACER-Bench. These scenarios are designed to probe basic yet challenging visual reasoning abilities of VLMs under cross-region dependency, where successful prediction requires not only recognizing local visual patterns but also maintaining and propagating visual evidence across spatially separated regions. 

\begin{table}[!htbp]
\centering
\small
\setlength{\tabcolsep}{3pt}
\caption{Scenario definitions of TRACER-Bench. ``RS Subimage Matching'' denotes Remote-Sensing Subimage Matching.}
\label{tab:tracer_bench_scenarios}
\begin{tabular}{p{0.25\linewidth} p{0.50\linewidth} p{0.18\linewidth}}
\toprule
Scenario & Required evidence & Answer type \\
\midrule
Rule-Switching Navigation
& Node attributes, directed edges, and rule switches 
& Final target node \\

Nonsemantic Tracing 
& Path identity under distractors 
& Endpoint label \\

Embedded Matching 
& Fine-grained template-grid correspondence 
& Binary decision \\

RS Subimage Matching
& Local image patch, scale, and rotation consistency 
& Binary decision \\
\bottomrule
\end{tabular}
\vspace{0mm}
\end{table}

\begin{figure}[!htbp]
    \centering
    \includegraphics[width=0.96\linewidth]{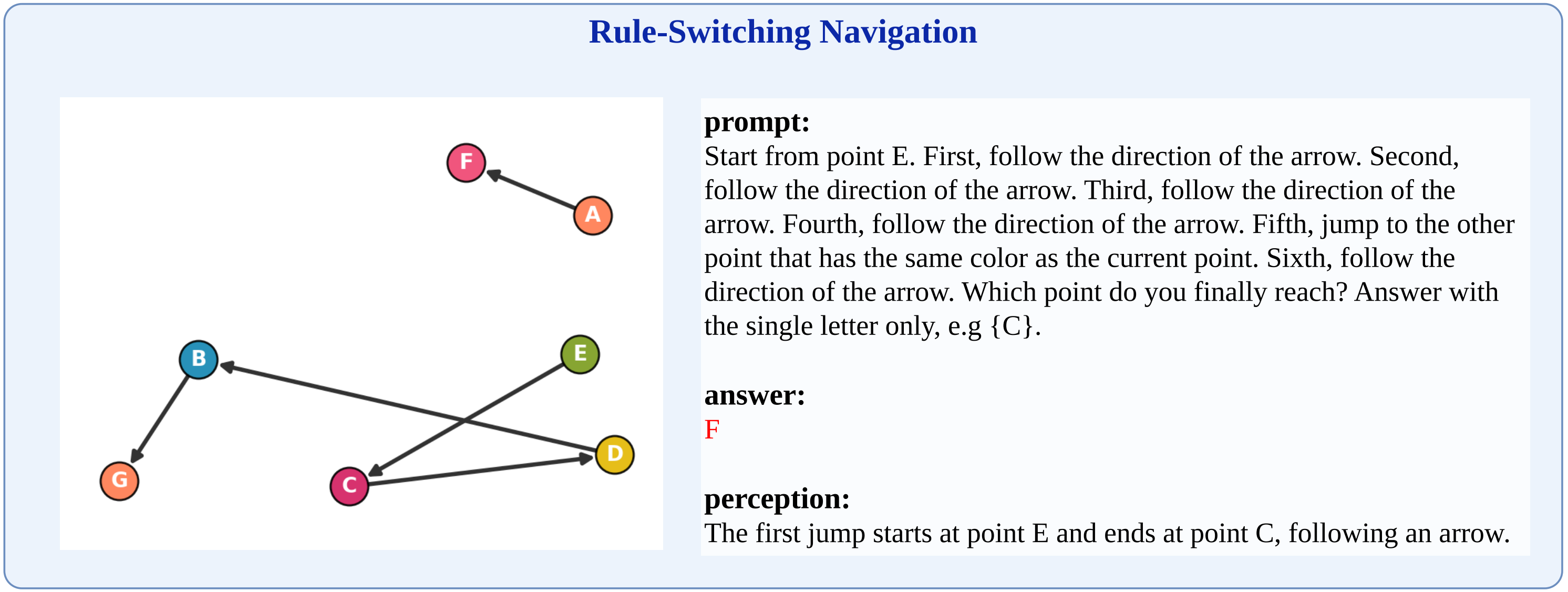}
    \caption{
    Representative example of Rule-Switching Navigation.
    }
    \label{fig:sample_rule_switching}
    \vspace{0mm}
\end{figure}

Figure~\ref{fig:sample_rule_switching} shows a sample of Rule-Switching Navigation. The model must maintain the current node as an intermediate state and update it according to the rule specified at each step. Because arrow-following and color-matching rules appear in an interleaved manner, the next target region may depend on either local directional cues or nonlocal attribute matching. This makes the evidence chain both spatially distributed and easily confusable: the model has to distinguish valid arrows from nearby distractors, compare colors across separated nodes, and propagate the current state through multiple transitions. Thus, the task tests whether VLMs can perform rule-conditioned state tracking and nonlocal evidence integration, rather than relying on single-region recognition.

\begin{figure}[!htbp]
    \centering
    \includegraphics[width=0.96\linewidth]{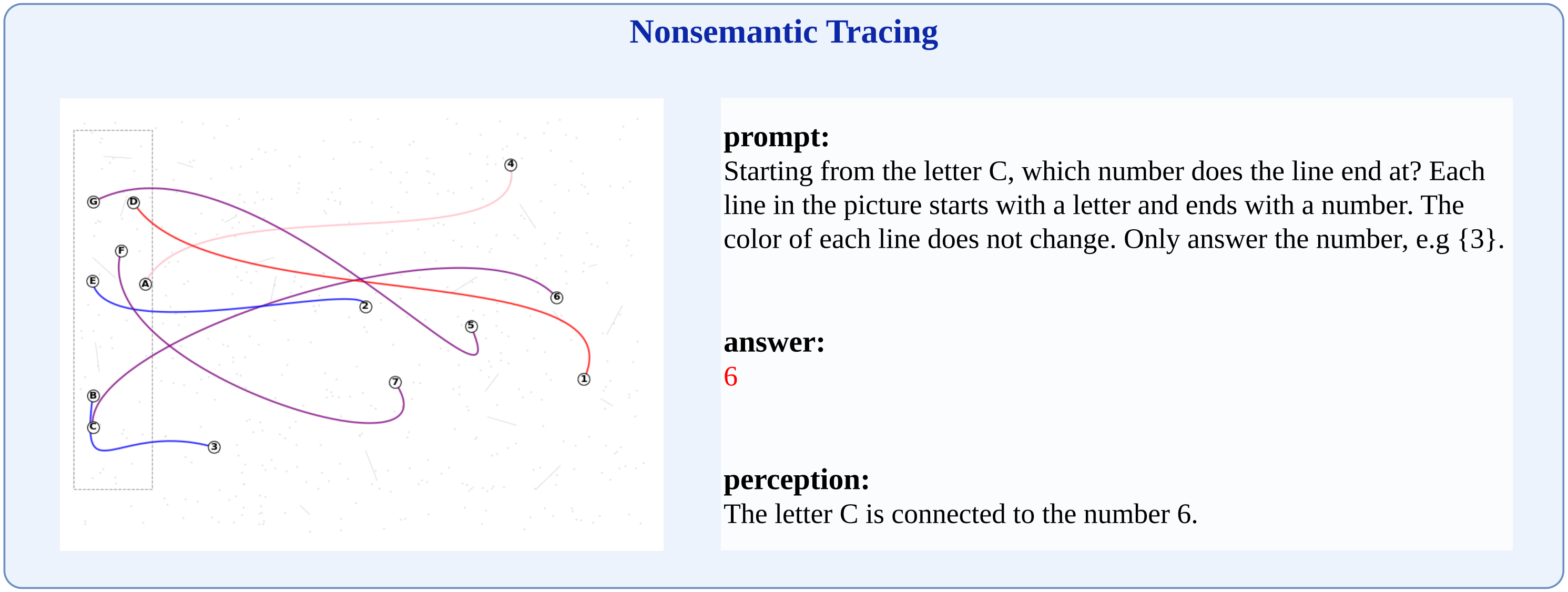}
    \caption{
    Representative example of Nonsemantic Tracing.
    }
    \label{fig:sample_nonsemantic_tracing}
    \vspace{0mm}
\end{figure}

Figure~\ref{fig:sample_nonsemantic_tracing} shows a sample of Nonsemantic Tracing. The model must follow a continuous curve from a specified letter endpoint to its corresponding numbered endpoint, without relying on semantic cues such as components, ports, or object identities. The task is inherently nonlocal because the correct endpoint is often spatially distant from the starting letter, and the model has to maintain path identity across a long curved trajectory. At the same time, the curve is easily confusable: multiple colored paths may overlap, run in parallel, or pass through nearby regions, while background noise introduces additional visual distractions. The curves are procedurally generated with randomized Bezier control points, which produces smooth but variable direction changes and prevents the model from relying on simple straight-line extrapolation. Thus, the task tests whether VLMs can perform continuous visual tracking under distractors, rather than solving the problem through local endpoint recognition or semantic priors.

\begin{figure}[!htbp]
    \centering
    \includegraphics[width=0.96\linewidth]{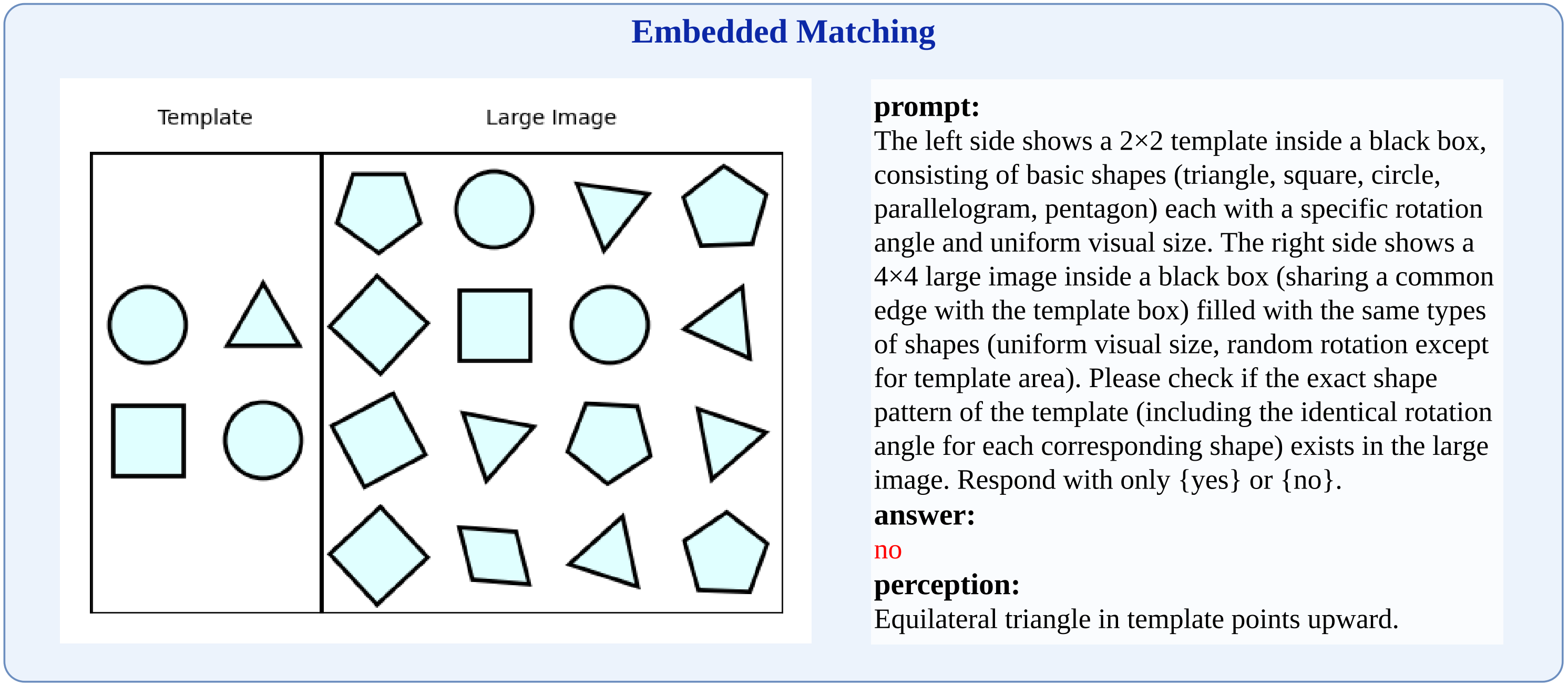}
    \caption{
    Representative example of Embedded Matching.
    }
    \label{fig:sample_embedded_matching}
    \vspace{0mm}
\end{figure}

Figure~\ref{fig:sample_embedded_matching} shows a sample of Embedded Matching. The model must determine whether the local template appears exactly within the larger structure, requiring simultaneous comparison of multiple shapes, their relative arrangement, and their rotations. This setting is easily confusable because rotated shapes may appear visually similar while still differing in orientation or local configuration, and such differences are difficult to describe completely within a bounded text representation. Moreover, matching multiple elements requires the model to retain previous comparison states while checking subsequent positions in the larger image. Thus, the task tests confusable local-to-global structural correspondence beyond single-shape recognition.

\begin{figure}[!htbp]
    \centering
    \includegraphics[width=0.96\linewidth]{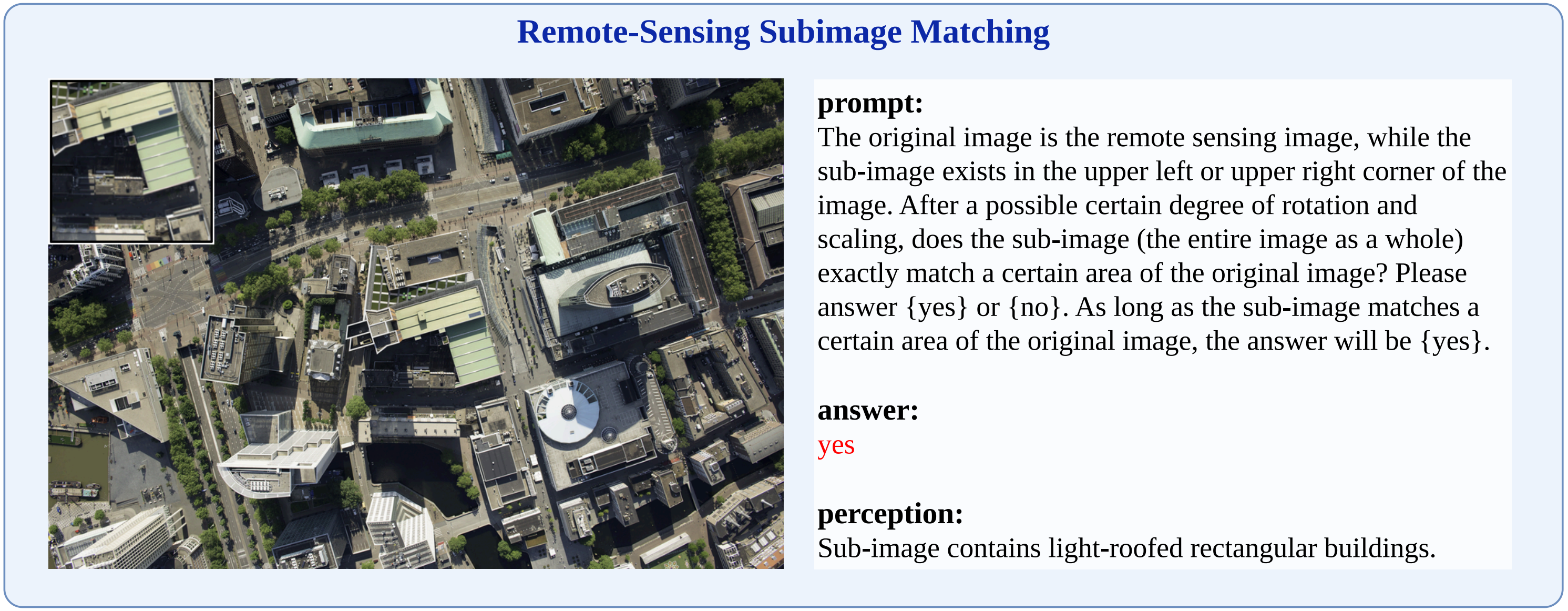}
    \caption{
    Representative example of Remote-Sensing Subimage Matching.
    }
    \label{fig:sample_rs_subimage_matching}
    \vspace{5mm}
\end{figure}

Figure~\ref{fig:sample_rs_subimage_matching} shows a sample of Remote-Sensing Subimage Matching. We derive the original remote-sensing images from MME-RealWorld~\citep{zhang2025mmerealworld}, while constructing new subimage-matching instances rather than reusing its original QA annotations. Compared with abstract scenes, these images contain richer visual content, including dense buildings, roads, shadows, and background clutter. The model must determine whether the candidate subimage corresponds to a region in the original image, even when rotation or scale changes are applied. These transformations make the evidence more confusable, since visually similar urban structures may appear at different orientations or resolutions. Thus, the task tests whether VLMs can maintain robust local-to-global correspondence in realistic scenes with complex background variation.

\subsection{Dataset statistics}

Table~\ref{tab:tracer_bench_stats} reports the dataset statistics of TRACER-Bench. All scenarios are organized with the same annotation interface. For binary matching tasks, we balance positive and negative samples. For chain-based and path-based tasks, we stratify the samples according to the corresponding difficulty factors, such as the number of nodes, path complexity, or distractor density.

\begin{table}[!htbp]
\centering
\small
\setlength{\tabcolsep}{3pt}
\caption{Statistics of TRACER-Bench.}
\label{tab:tracer_bench_stats}
\begin{tabular}{
p{0.28\linewidth}
p{0.1\linewidth}
p{0.12\linewidth}
p{0.39\linewidth}
}
\toprule
Scenario & \#Samples & Answer type & Main difficulty factors  \\
\midrule
Rule-Switching Navigation 
& 245 
& Multi-class 
& Nodes / transitions / distractors  \\

Nonsemantic Tracing 
& 245 
& Multi-class 
& Number of lines / path ambiguity  \\

Embedded Matching 
& 245 
& Binary 
& Candidate size / structural distractors  \\

RS Subimage Matching
& 245 
& Binary 
& Scale / rotation / background complexity  \\
\midrule
Total & 980 & -- & --  \\
\bottomrule
\end{tabular}
\vspace{1mm}
\begin{minipage}{0.95\linewidth}
\footnotesize
\textbf{Note:} ``RS Subimage Matching'' denotes Remote-Sensing Subimage Matching.
\end{minipage}
\end{table}

\paragraph{Data format}

Each TRACER-Bench sample is stored as one JSONL entry. Table~\ref{tab:tracer_bench_format} describes the unified field format. The \texttt{perception} field is not intended to be a free-form explanation. Instead, it records task-critical intermediate visual evidence derived from the generation process or annotation procedure, and is used only for process-level analysis or supervision. A representative JSONL example is shown as listing~\ref{lst:tracer_bench_example}.

\begin{table}[t]
\centering
\small
\caption{Unified sample format of TRACER-Bench.}
\label{tab:tracer_bench_format}
\begin{tabular}{p{0.18\linewidth} p{0.72\linewidth}}
\toprule
Field & Description \\
\midrule
\texttt{id} & Unique sample identifier. \\
\texttt{image} & Relative path to the input image. \\
\texttt{prompt} & Natural-language instruction describing the visual reasoning task. \\
\texttt{answer} & Final target label or binary decision. \\
\texttt{perception} & Key intermediate visual evidence required to solve the task. \\
\bottomrule
\end{tabular}
\end{table}

\begin{lstlisting}[
basicstyle=\ttfamily\scriptsize,
breaklines=true,
frame=single,
caption={Example JSONL entry from the Embedded Matching scenario.},
label={lst:tracer_bench_example}
]
{
  "id": "match_0",
  "image": "./images/exp_0000.png",
  "prompt": "The left side shows a 2x2 template inside a black box, consisting of basic shapes (triangle, square, circle, parallelogram, pentagon) each with a specific rotation angle and uniform visual size. The right side shows a 4x4 large image inside a black box (sharing a common edge with the template box) filled with the same types of shapes (uniform visual size, random rotation except for template area). Please check if the exact shape pattern of the template (including the identical rotation angle for each corresponding shape) exists in the large image. Respond with only {yes} or {no}.",
  "answer": "yes",
  "perception": "Equilateral triangle in template points upward."
}
\end{lstlisting}

\paragraph{Difficulty factors}

TRACER-Bench controls difficulty through task-specific factors that correspond to different forms of cross-region dependency. Table~\ref{tab:tracer_bench_difficulty} summarizes the main factors used in each scenario. These factors are not intended to define a single universal difficulty score. Instead, they provide interpretable axes for analyzing how model performance changes under longer dependency chains, denser visual distractors, or more complex local-to-global correspondence.

\begin{table}[t]
\centering
\small
\setlength{\tabcolsep}{3pt}
\caption{Difficulty factors used in TRACER-Bench.}
\label{tab:tracer_bench_difficulty}
\begin{tabular}{p{0.24\linewidth} p{0.53\linewidth} p{0.17\linewidth}}
\toprule
Difficulty factor & Definition & Applied scenarios \\
\midrule
Dependency length 
& Number of required reasoning hops or state updates. 
& VJump \\

Structural distractors 
& Number of irrelevant shapes or candidate substructures. 
& Match \\

Cross-region distance 
& Spatial span between key evidence regions. 
& VJump, LT \\

Scale/rotation variation 
& Scale or rotation change between the candidate and source patch. 
& Match, MME-RS \\
\bottomrule
\end{tabular}
\vspace{1mm}
\begin{minipage}{0.95\linewidth}
\footnotesize
\end{minipage}
\end{table}

\subsection{Quality control}
\label{app:quality_control}

For synthetic scenarios, final answers and perception fields are derived from the underlying generation states, such as sampled paths, template configurations, endpoint mappings, or patch correspondences. To ensure annotation reliability, the generated samples were manually inspected by the authors to check the consistency among visual content, prompts, final answers, and perception fields. For binary matching tasks, we maintain a balanced positive/negative distribution. For three abstract scenarios, we perform stratified sampling across difficulty levels, such as dependency length, path complexity, candidate size, and visual distractor density. This prevents the benchmark from being dominated by overly simple or overly difficult cases, and yields a more balanced evaluation of model performance across different reasoning conditions. For remote-sensing subimage matching, candidate patches are selected to preserve sufficient visual information and avoid overly confusable or low-texture crops. Samples with unclear correspondences, severe visual artifacts, or inconsistent annotations are removed during inspection. 

The perception field is designed as a compact description of task-critical visual evidence rather than a free-form rationale. Because the target tasks are precisely those where current VLMs are brittle, using AI-generated rationales to annotate evidence would make the evidence reliability questionable and substantially increase the burden of manual verification. Instead, we derive perception units from the latent generation state of each scripted scenario, which allows us to control the evidence type, difficulty, and dependency structure during data construction. We keep only evidence necessary for solving the instance, such as local attributes, path transitions, rule-dependent correspondences, endpoint relations, or patch-level matches, and avoid redundant explanations or direct restatements of the final answer. During inspection, we check whether each perception unit is visually grounded, necessary for the answer, and consistent with the prompt and generated image. The perception field is used only for computing process rewards and offline analysis, and is never provided to the model during evaluation.

\section{Training data construction and split protocol}
\label{app:training_data_construction}

This section provides additional details on the training data used for \modelname reinforcement learning. The training corpus shares the same scenario families with \datasetname, but is generated independently from the benchmark. In addition, we construct non-overlapping Tunnel-Vision-style synthetic data as auxiliary training data and convert it into the unified \textbf{prompt-answer-perception} format.

\subsection{Training corpus composition}
\label{app:training_corpus_composition}

\begin{figure}[t]
    \centering
    \includegraphics[width=0.6\linewidth]{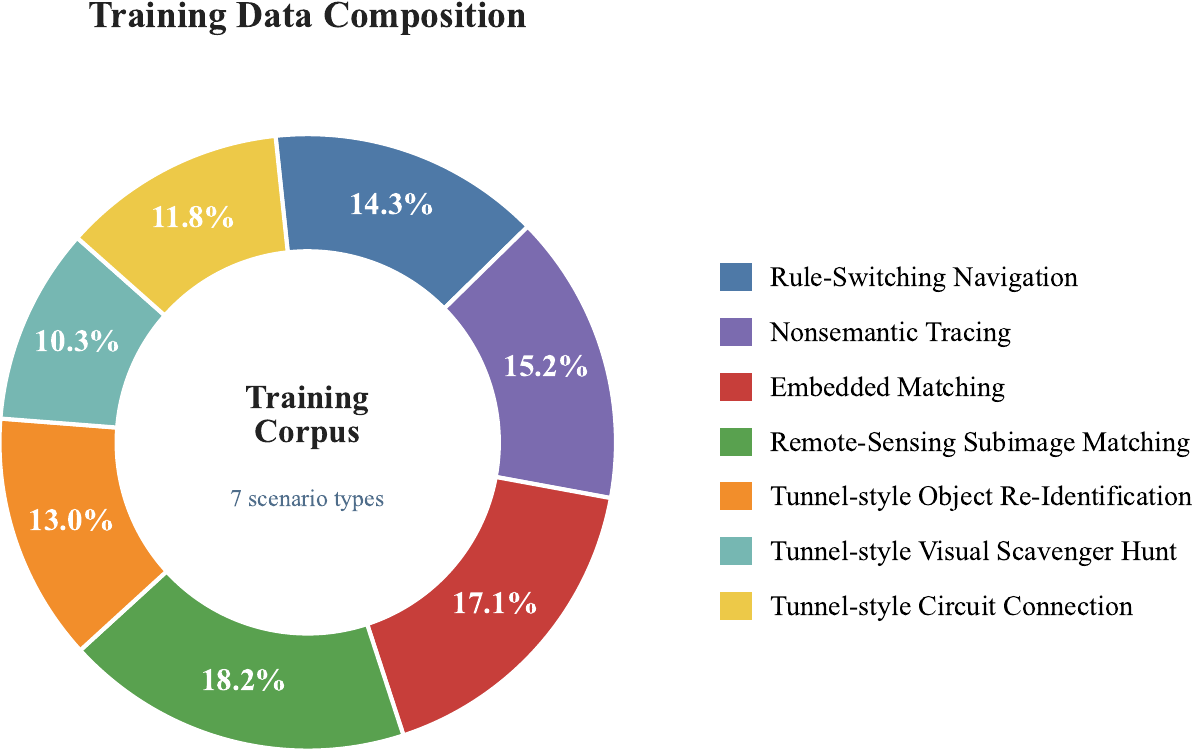}
    \caption{
    Composition of the training corpus used for \modelname reinforcement learning. The corpus consists of four TRACER-style scenario families and three independently generated Tunnel-Vision-style auxiliary scenario types.
    }
    \label{fig:training_data_composition}
\end{figure}

Figure~\ref{fig:training_data_composition} summarizes the composition of the training corpus, which combines four TRACER-style scenario families with three Tunnel-Vision-style auxiliary scenario types. The auxiliary Tunnel-Vision-style data is used only for training, is not part of \datasetname, and does not overlap with the public Tunnel Vision evaluation set.

\subsection{Split and isolation protocol}
\label{app:split_isolation_protocol}

To prevent instance leakage, we separate the training corpus and \datasetname at both the sample and generator-configuration levels, as summarized in Table~\ref{tab:split_isolation_protocol}. This protocol reduces the risk of instance memorization and encourages evaluation on unseen layouts, image regions, and dependency configurations.

\begin{table}[t]
\centering
\small
\caption{Split and isolation protocol between the training corpus and \datasetname.}
\label{tab:split_isolation_protocol}
\begin{tabular}{p{0.26\linewidth} p{0.66\linewidth}}
\toprule
\textbf{Isolation factor} & \textbf{Protocol} \\
\midrule
Random seeds 
& Training and benchmark samples are generated with disjoint random seeds. \\

Generated layouts 
& Exact generated layouts are not reused between the training corpus and \datasetname. \\

Rule configurations 
& Training and benchmark data may share abstract rule types, while concrete ordered rule sequences are independently sampled rather than copied across splits. \\

Image regions 
& For real-image scenarios, training and benchmark patches are sampled from disjoint images or non-overlapping regions. \\

Concrete instances 
& No identical image--question--answer tuple appears in both the training corpus and \datasetname. \\

Shared difficulty axes 
& Training and benchmark data may share difficulty axes, such as dependency length, cross-region distance, and distractor level, but not concrete generated configurations. \\

Tunnel-Vision-style data
& Auxiliary Tunnel-Vision-style training data is generated independently and does not overlap with the public Tunnel Vision evaluation set~\citep{berman2025tunnelvision}. \\
\bottomrule
\end{tabular}
\vspace{5pt}
\end{table}

\subsection{Tunnel-Vision-style example and perception}
\label{app:Tunnel-Vision-style_example}

\begin{figure}[t]
    \centering
    \includegraphics[width=1\linewidth]{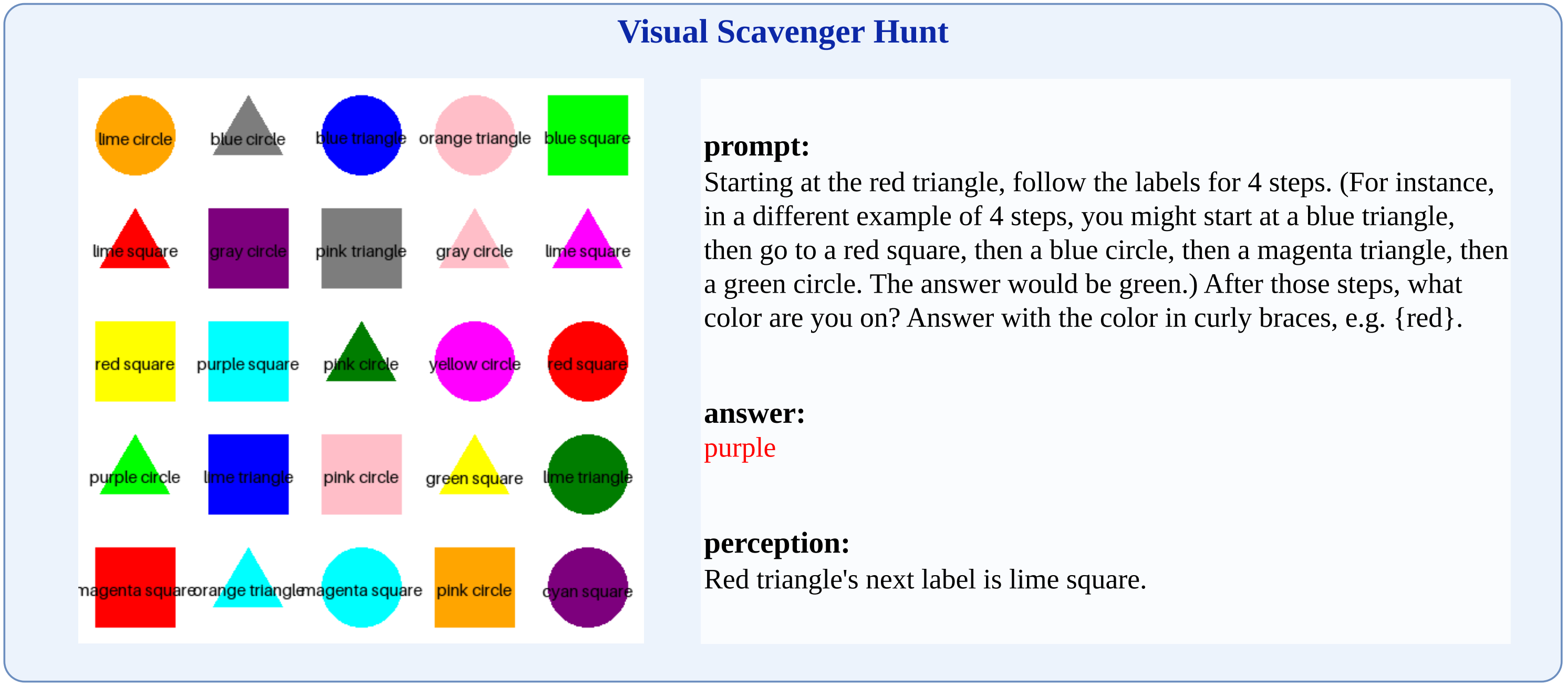}
    \caption{
    Representative Tunnel-Vision-style jumping example in the training corpus. The sample is converted into the unified prompt-answer-perception format.
    }
    \label{fig:Tunnel-Vision-style_jump_example}
    \vspace{15pt}
\end{figure}

Figure~\ref{fig:Tunnel-Vision-style_jump_example} shows a representative Tunnel-Vision-style jumping example used in the training corpus. The task requires the model to follow a sequence of visual transitions across spatially separated regions. To unify such auxiliary data with the main Tracer-style training scenarios, we supplement each sample with a \texttt{perception} field.

The \texttt{perception} field is constructed from the generation process rather than written as a free-form rationale. For Tunnel-Vision-style jumping scenes, it records the minimal visual evidence needed to support the region transitions, such as the start region, the visual cue used for each jump, and the reached target region. We deliberately keep this field concise: it should describe only the key visual evidence required for supervision, rather than provide a long chain-of-thought explanation. This design allows the auxiliary synthetic data to share the same training interface as the main scenarios while providing a verifiable intermediate signal for process-level optimization.

\section{Train progress}
\subsection{Training and reward hyperparameters}
\label{app:training_details}

\paragraph{Training details}
CAVE uses Qwen2.5-VL as the base model and adopts a two-stage training procedure on 8 A800 GPUs \citep{bai2025qwen25vl,sheng2024hybridflow,shao2024deepseekmath}. First, we use 10k DeepEyes-style samples for cold-start training, enabling the model to acquire basic interleaved reasoning and tool-use capabilities \citep{zheng2025deepeyes}. Then, we perform process-reward training on 5,000 TRACER-style samples with process evidence annotations \citep{shao2024deepseekmath,berman2025tunnelvision}, and incorporate constraints on answer correctness, format validity, and the number of rounds to stabilize optimization. CAVE-3B adopts the same training procedure but replaces the backbone with a 3B model; Belief-Only keeps the same backbone, data, and optimization settings, while replacing the structured process rewards with answer-belief gain only. Tool observations are used only as contextual feedback, with their response masks set to $0$, and do not participate in policy-gradient computation. The weights of the three structured credits are set to a ratio of $\lambda_{\mathrm{bu}}:\lambda_{\mathrm{ea}}:\lambda_{\mathrm{af}}=4:3:3$, selected based on validation performance after a small hyperparameter sweep.

\paragraph{Reward aggregation.}
CAVE combines three structured process credits: belief update, evidence acquisition, and adaptive focus control. For each rollout trajectory $\tau$, the final structured process reward is computed as
\[
R_{\mathrm{CAVE}}(\tau)
=
\lambda_{\mathrm{bu}} C_{\mathrm{bu}}(\tau)
+
\lambda_{\mathrm{ea}} C_{\mathrm{ea}}(\tau)
+
\lambda_{\mathrm{af}} C_{\mathrm{af}}(\tau),
\]
where $C_{\mathrm{bu}}$, $C_{\mathrm{ea}}$, and $C_{\mathrm{af}}$ denote the aggregated credits for answer-belief update, task-critical evidence acquisition, and uncertainty-aware adaptive focus control, respectively. In all main experiments, we use
\[
\lambda_{\mathrm{bu}}:\lambda_{\mathrm{ea}}:\lambda_{\mathrm{af}}=4:3:3 .
\]
Equivalently, the normalized weights are $0.4$, $0.3$, and $0.3$.

\paragraph{Selection protocol.}
We select the reward weights using a small validation-based sweep on the validation split constructed from the training sources, without using any reported test benchmark for hyperparameter selection. We consider several nearby configurations around balanced weighting, including equal weighting and belief-dominant variants. The final ratio $\lambda_{\mathrm{bu}}:\lambda_{\mathrm{ea}}:\lambda_{\mathrm{af}}=4:3:3$ gives slightly higher weight to belief update, which provides the most stable answer-side progress signal, while retaining substantial supervision from evidence acquisition and adaptive focus control. This avoids reducing the reward to a single answer-side information-gain objective and improves early-stage RL stability. In all experiments, we set the entropy top-$k$ to $k=500$, fix the crop-scale range to $\rho_{\min}=0.02$ and $\rho_{\max}=0.30$, use a round-decay weight $w_t^m=0.8^{t-1}$, apply $\phi_m(c)=\mathrm{clip}(c,-1,2)$ to bound credit magnitudes, and limit each rollout to at most $T=5$ reasoning rounds.

\paragraph{Use across experiments.}
We use the same reward and rollout settings for the final CAVE model and the smaller-backbone CAVE-3B variant. For reward ablations, the corresponding credit term is removed while the remaining terms keep their relative proportions. The Belief-Only baseline keeps the same backbone, cold-start initialization, training data, rollout setting, and optimization recipe, but replaces the structured reward with the belief-update credit alone. Thus, differences between CAVE and its controlled variants mainly reflect the effect of structured credit composition rather than changes in training setup.
\vspace{-3mm}

\paragraph{Stability considerations.}
During training, we monitor validation reward, answer accuracy, and KL divergence to avoid reward-scale imbalance or policy drift. The selected weighting yields stable reward improvement without observed policy collapse. Since GRPO uses group-relative advantages, the absolute reward scale is less important than the relative ranking of trajectories within each rollout group; nevertheless, we keep the same reward normalization and clipping strategy for all compared variants to ensure fair comparison.

\subsection{Experimental resource consumption}
\label{app:resource_consumption}

All training experiments are conducted on a single node with 8 NVIDIA A800 80GB GPUs. 
We implement the reinforcement learning stage with the verl framework and use mixed-precision training and gradient checkpointing to reduce memory consumption. 
For CAVE-7B, the cold-start supervised fine-tuning stage on 10K interleaved reasoning samples takes approximately 10 hours. The subsequent CAVE process-reward training stage on 5K samples takes approximately 24 hours, including policy rollout, tool-based visual re-observation, structured credit computation, and GRPO optimization.

\subsection{Process behavior and efficiency}
\label{app:budget_controlled}

To verify that the improvement of CAVE does not simply come from longer responses,
more reasoning rounds, or more frequent tool use, we compare the inference budget
and process behavior of DeepEyes, the cold-start model, and CAVE under the same
evaluation protocol. Specifically, we report the average accuracy, number of
assistant-generated tokens, reasoning rounds, zoom-tool calls, and inference time
per sample. Tool-returned observations are treated as environment feedback and
are not counted as policy-generated tokens.

\begin{figure*}[t]
    \centering
    \begin{subfigure}[t]{0.24\textwidth}
        \centering
        \caption{Total reward.}
        \includegraphics[width=\linewidth]{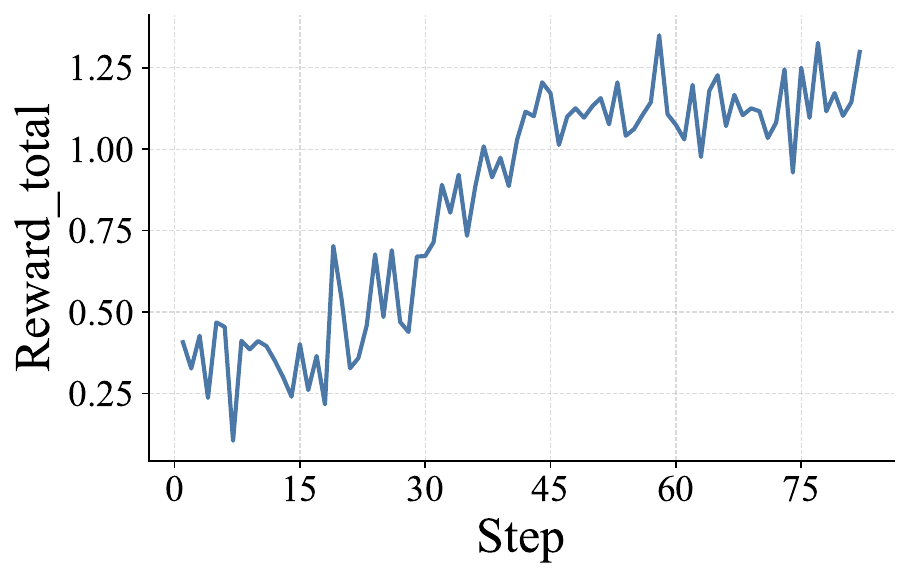}
        \label{fig:reward_total_curve}
    \end{subfigure}
    \hfill
    \begin{subfigure}[t]{0.24\textwidth}
        \centering
        \caption{Belief reward.}
        \includegraphics[width=\linewidth]{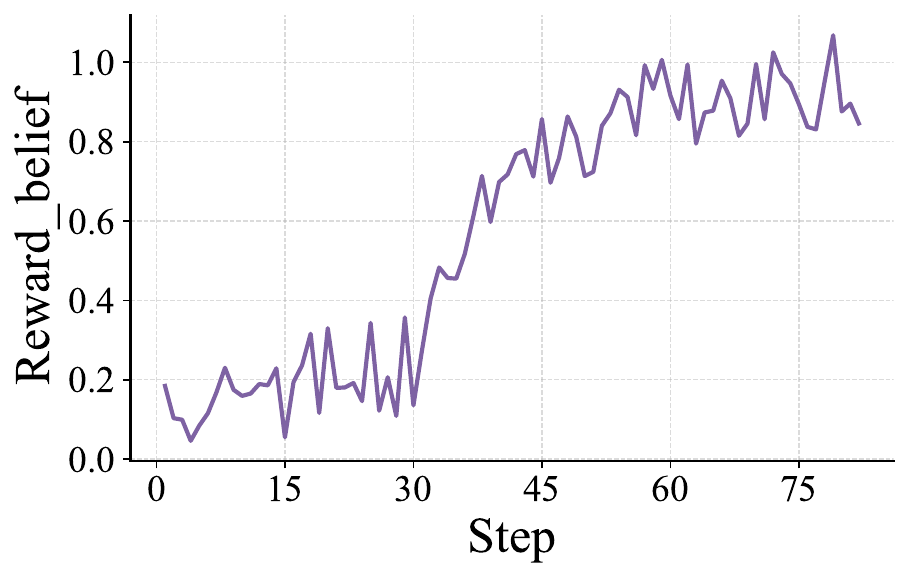}
        \label{fig:reward_belief_curve}
    \end{subfigure}
    \hfill
    \begin{subfigure}[t]{0.24\textwidth}
        \centering
        \caption{Evidence reward.}
        \includegraphics[width=\linewidth]{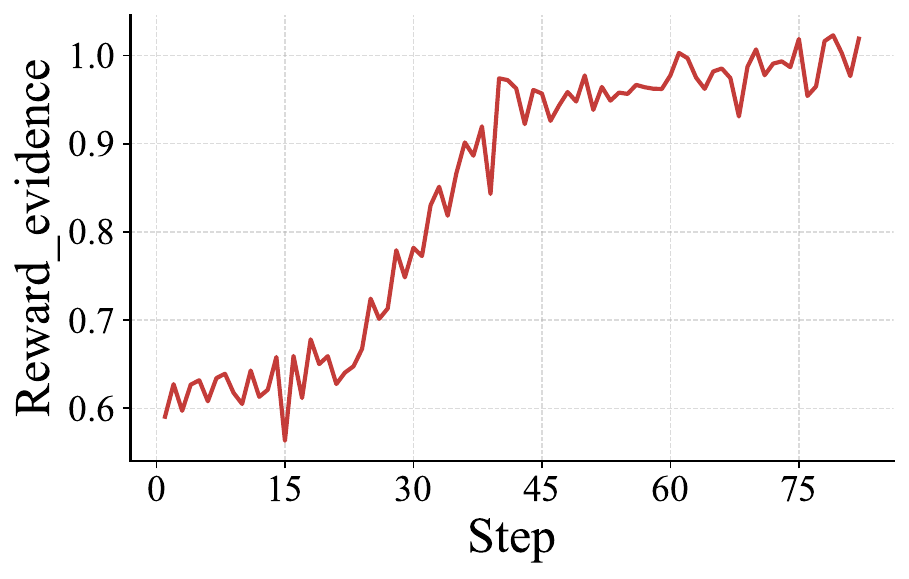}
        \label{fig:reward_evidence_curve}
    \end{subfigure}
    \hfill
    \begin{subfigure}[t]{0.24\textwidth}
        \centering
        \caption{Focusing reward.}
        \includegraphics[width=\linewidth]{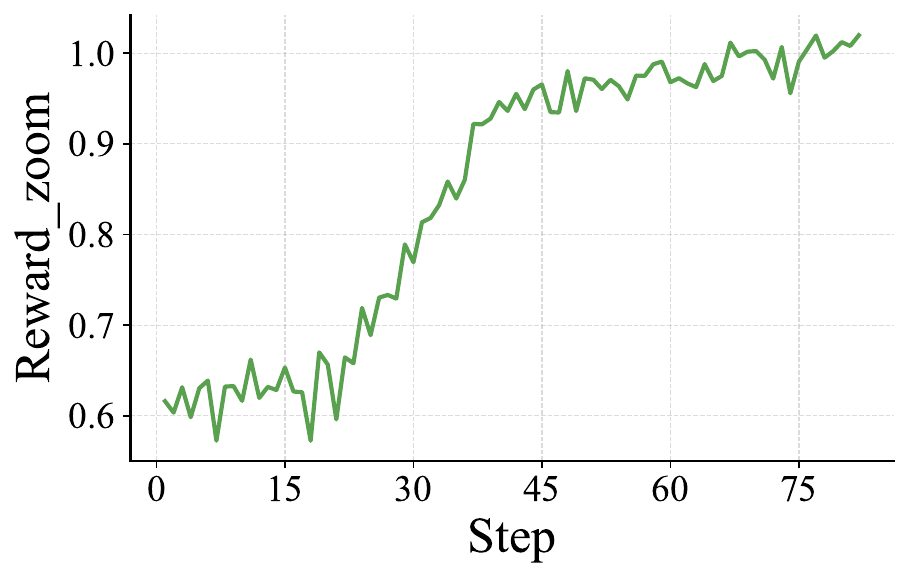}
        \label{fig:reward_zoom_curve}
    \end{subfigure}
    \vspace{-10pt}
    \caption{
    Training dynamics of CAVE process rewards. 
    The total structured reward and its three components---belief update, evidence acquisition, and adaptive focus control---improve steadily, indicating that CAVE effectively optimizes the intended visual evidence progression signals.
    }
    \label{fig:training_reward_dynamics}
\vspace{0pt}
\end{figure*}

\begin{figure*}[t]
    \centering
    \includegraphics[width=0.48\textwidth]{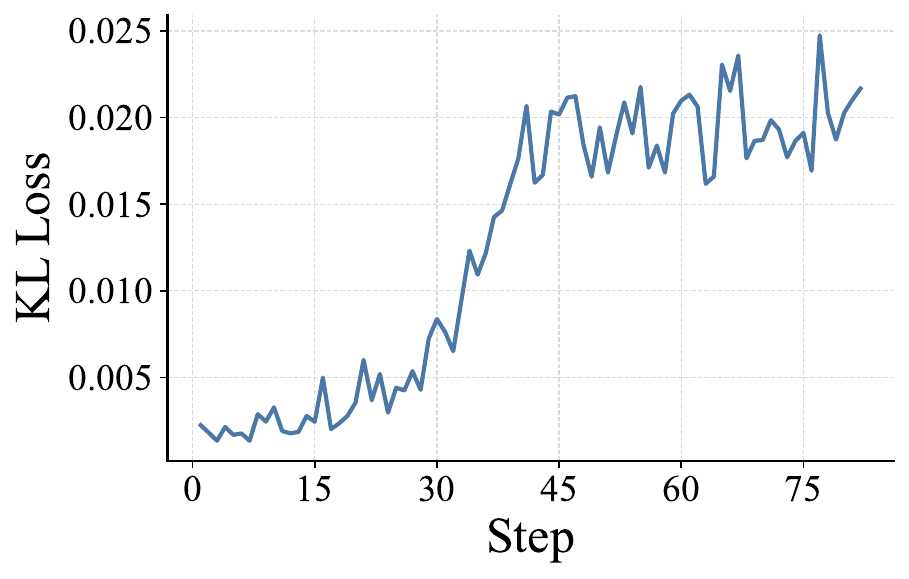}
    \caption{
    KL dynamics during CAVE training. 
    The KL loss increases in the early stage and then remains within a controlled range, suggesting stable optimization without obvious policy collapse or abnormal drift.
    }
    \label{fig:training_kl_curve}
\vspace{2pt}
\end{figure*}

As shown in Table~\ref{tab:budget_controlled}, CAVE achieves higher accuracy without
requiring a substantially larger inference budget. Compared with the cold-start
model, CAVE improves the effectiveness of intermediate reasoning and visual
evidence acquisition rather than merely increasing the number of rounds or tool
calls. Compared with DeepEyes, CAVE maintains a comparable tool-use budget while
achieving stronger performance, suggesting that structured process credits help
the model select and exploit task-relevant visual evidence more efficiently.

\section{Teacher forcing details}
\label{app:teacher_forcing}

\begin{figure}[t]
\centering
\includegraphics[width=\linewidth]{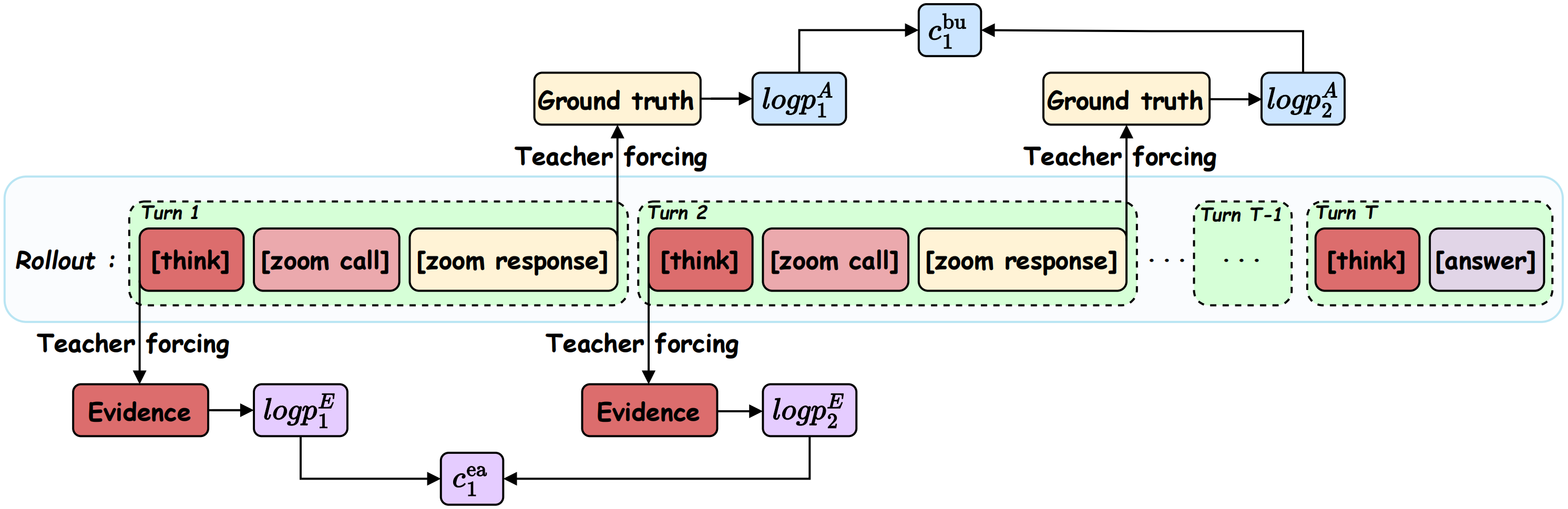}
\caption{Teacher-forced answer and evidence gains.}
\label{fig:teacher_forcing_gain}
\vspace{5pt}
\end{figure}

Figure~\ref{fig:teacher_forcing_gain} illustrates how teacher forcing is used to compute answer-belief gain and evidence gain. After a rollout is generated, we separately score fixed target sequences under each context state. For belief update, the target is the ground-truth answer body; for evidence acquisition, the targets are textualized evidence units. The adjacent-state log-likelihood gain from $\mathcal M_t$ to $\mathcal M_{t+1}$ is used as the corresponding process credit. This scoring pass is only used for reward computation and does not affect rollout generation or later context states.

\section{Statistical significance}
\label{app:statistical_significance}

We conduct a statistical significance analysis for the main same-backbone comparison between CAVE-7B and Qwen2.5-VL$^\ast$. For each benchmark group, we model accuracy as a binomial proportion over test instances and report 95\% confidence intervals. Since the two models are evaluated on the same test instances, we further apply McNemar's test to the paired correct/incorrect outcomes. As shown in Table~\ref{tab:statistical_significance}, the observed gains remain statistically meaningful at the aggregate benchmark level, supporting the robustness of the same-backbone improvements.

\begin{table}[!htbp]
\centering
\small
\setlength{\tabcolsep}{4.5pt}
\renewcommand{\arraystretch}{1.12}
\caption{
Statistical significance analysis for the main same-backbone comparison between CAVE-7B and Qwen2.5-VL$^\ast$. Bracketed values report 95\% normal-approximation confidence intervals for aggregate accuracies and accuracy gaps, and $p$-values are obtained using McNemar's test on paired correct/incorrect outcomes.
}
\label{tab:statistical_significance}
\begin{tabular*}{\linewidth}{@{\extracolsep{\fill}}lccccc@{}}
\toprule
Benchmark & $N$ & CAVE-7B & Qwen2.5-VL$^\ast$ & $\Delta$ & $p$-value \\
\midrule
Tunnel Vision 
& 1350 
& 31.3 [28.9, 33.8] 
& 24.4 [22.1, 26.7] 
& +7.0 [3.6, 10.3] 
& $5.4{\times}10^{-5}$ \\
BabyVision 
& 388 
& 17.3 [13.5, 21.1] 
& 11.0 [7.8, 14.1] 
& +6.3 [1.4, 11.2] 
& $1.23{\times}10^{-2}$ \\
TRACER-Bench 
& 980 
& 42.5 [39.4, 45.5] 
& 34.0 [31.0, 37.0] 
& +8.5 [4.2, 12.7] 
& $1.18{\times}10^{-4}$ \\
\bottomrule
\end{tabular*}
\end{table}

\section{Limitations}
\label{app:limitations}

Our empirical evaluation focuses on open-source VLM backbones at moderate model scales, allowing controlled same-backbone comparisons and detailed analysis of intermediate reasoning behaviors. While CAVE is not architecture-specific and only assumes an autoregressive interleaved reasoning interface, its scaling behavior on substantially larger VLMs remains an interesting direction for future investigation. Moreover, the current implementation uses localized zoom-based re-observation as the main visual action space. Future work may extend structured credit assignment to richer multimodal actions, such as explicit region comparison, object-level grounding, and external visual tool invocation.

\section{Broader impacts}
\label{app:broader_impacts}

CAVE is intended to improve the reliability and interpretability of VLMs by encouraging models to acquire and integrate task-relevant visual evidence during interleaved reasoning. Together with TRACER-Bench, it provides a controlled framework for diagnosing whether models rely on grounded visual evidence rather than superficial language priors. These properties may benefit the development of more robust multimodal systems.

However, stronger visual evidence exploration may also be misused in sensitive visual analysis scenarios, such as surveillance or fine-grained monitoring. Moreover, performance gains on controlled benchmarks should not be taken as sufficient evidence for safe deployment in high-stakes applications. We therefore position this work as a research contribution to evidence-grounded visual reasoning, and encourage task-specific validation, human oversight, and appropriate safeguards before any real-world use.

\section{Existing assets and licenses}
\label{app:existing_assets_licenses}

This work uses existing assets only for model initialization, baseline comparison, evaluation, and controlled data construction. We properly credit the original creators through citations to the corresponding papers, model cards, datasets, or repositories. We use all third-party assets under their released licenses or terms of use, and we do not claim ownership of any existing model, dataset, benchmark, or codebase.

Table~\ref{tab:existing_assets_licenses} summarizes the main existing assets used in this work. Public model checkpoints, including Qwen2.5-VL, Qwen3-VL, InternVL3.5, MiniCPM-V 4.5, R1-OneVision, ThinkLite-VL, and DeepEyes, are used only for initialization or baseline evaluation under their respective licenses. Existing benchmarks and datasets, including Tunnel Vision, BabyVision, MME-RealWorld, MMStar, RealWorldQA, HallusionBench, BLINK, MathVista, and MathVerse, are used only for research evaluation or controlled data construction. We do not redistribute third-party model weights or restricted benchmark assets as part of our released package. For derived or generated data constructed in this work, we release only the generated samples and metadata that are permitted by the corresponding source licenses and terms of use.

\begin{table}[t]
\centering
\small
\caption{Existing assets used in this work and their licenses or terms of use.}
\label{tab:existing_assets_licenses}
\begin{tabular}{p{0.25\linewidth} p{0.47\linewidth} p{0.18\linewidth}}
\toprule
\textbf{Asset} & \textbf{Use in this work} & \textbf{License / terms} \\
\midrule
Qwen2.5-VL & Base model and baseline & Apache-2.0 \\
Qwen3-VL & Baseline evaluation & Apache-2.0 \\
InternVL3.5 & Baseline evaluation & Apache-2.0 \\
MiniCPM-V 4.5 & Baseline evaluation & Apache-2.0 \\
R1-OneVision & Baseline evaluation & Apache-2.0 \\
ThinkLite-VL & Baseline evaluation & MIT \\
DeepEyes & Baseline/cold-start data reference & Apache-2.0 \\
Tunnel Vision & Benchmark and data reference & BSD-3-Clause \\
BabyVision & Public benchmark evaluation & MIT \\
MME-RealWorld & Remote-sensing source and evaluation & Apache-2.0 \\
MMStar & General multimodal benchmark evaluation & Apache License 2.0 \\
RealWorldQA & General multimodal benchmark evaluation & CC-BY-ND-4.0 \\
HallusionBench & General multimodal benchmark evaluation & BSD-3-Clause \\
BLINK & General multimodal benchmark evaluation & Apache-2.0 \\
MathVista & General multimodal benchmark evaluation & CC-BY-SA-4.0 \\
MathVerse & General multimodal benchmark evaluation & MIT \\
\bottomrule
\end{tabular}
\vspace{15pt}
\end{table}

\section{Additional characterization of fragmented visual evidence reasoning}
\label{app:fragmented_vision_formulation}

Given an image $I$ and a question $q$, we view a fragmented visual evidence reasoning instance as requiring a set of task-critical visual observations distributed across multiple regions. Let $G_q=(V_q,E_q)$ denote a query-induced dependency graph, where each node $v_i\in V_q$ corresponds to an image region $R_i\subset I$ and each edge $e_{ij}\in E_q$ indicates that evidence from region $R_i$ is needed to determine or verify evidence in region $R_j$. We denote the corresponding task-critical visual evidence units as
\[
\mathcal P_q = \{p_1,\ldots,p_K\},
\]
where each evidence unit $p_k$ is associated with a node, an edge, or a small subgraph in $G_q$, such as a local attribute, a path transition, or a cross-region correspondence. The answer can then be written abstractly as
\[
y = g(\mathcal P_q, q),
\]
where $\mathcal P_q$ denotes the visual evidence required by the question.

Such tasks are characterized by two properties. First, the required evidence is often nonlocal: solving the instance requires integrating information from spatially separated regions rather than recognizing a single local object. Second, the evidence is difficult to fully preserve through short textual descriptions. Fine-grained attributes, path continuity, local--global correspondences, and visually confusable differences may be lost or become confusable once compressed into language.

These properties make fragmented visual evidence reasoning naturally sequential. At each step, the model must decide which region to inspect, which evidence to retain, and how to update its current hypothesis based on previous observations. Therefore, the value of an intermediate action may not be immediately reflected in final answer correctness. A useful step may instead improve the model's answer belief, acquire a missing piece of task-critical evidence, or select a more appropriate region for the next observation. This motivates the structured credit assignment in CAVE, which decomposes intermediate progress into belief update, evidence acquisition, and adaptive focus control.

\end{document}